\documentclass[sigconf]{acmart}

\AtBeginDocument{%
  }

\usepackage{subcaption}
\usepackage{algorithm}
\usepackage{algorithmic}

\newcommand{\ours}{\texttt{GRILL}}
\newcommand{\oursName}{Gradient Signal Restoration in Ill-Conditioned Layers }

\newcommand{\oursMini}{\texttt{LGR}}
\newcommand{\oursMiniName}{Latent Gradient Restoration }

\renewcommand\footnotetextcopyrightpermission[1]{}

\setcopyright{none}

\renewcommand\footnotetextcopyrightpermission[1]{}
\setcopyright{none}
\begin{document}

\title{Revealing Hidden Vulnerabilities in Autoencoders through Gradient Signal Restoration}

\author{Chethan Krishnamurthy Ramanaik}
\author{Arjun Roy}
\author{Tobias Callies}
\author{Eirini Ntoutsi}

\affiliation{%
  \institution{University of the Bundeswehr Munich}
  \country{Germany}
}

\email{{chethan.krishnamurthy, arjun.roy, tobias.callies, eirini.ntoutsi}@unibw.de}

\begin{abstract}
Adversarial robustness of deep autoencoders (AEs) has received less attention than that of discriminative models, although their compressed latent representations induce ill-conditioned mappings that can amplify small input perturbations and destabilize reconstructions. Existing white-box attacks for AEs, which optimize norm-bounded adversarial perturbations to maximize reconstruction damage, often converge to suboptimal perturbations, thereby potentially overstating AE robustness. We show that this limitation is linked to vanishing adversarial loss gradients during backpropagation through ill-conditioned layers, associated with near-zero singular values in their intermediate weight matrices. To address this, we propose \ours{} (Gradient Signal Restoration in Ill-Conditioned Layers), a framework designed to mitigate gradient degradation and improve the reliability of adversarial robustness evaluation in encoder--decoder architectures. \ours{} is designed to mitigate adversarial gradient degradation during optimization, enabling attacks to better approximate high-distortion perturbations under fixed norm constraints. Through extensive experiments across multiple AE architectures, under both sample-specific and universal attacks, as well as standard and adaptive attack settings, we show that \ours{} significantly increases attack effectiveness, thereby exposing vulnerabilities hidden by existing attack limitations. Beyond AEs, we provide preliminary evidence that modern multimodal encoder–decoder architectures exhibit similar vulnerabilities.

    \href{https://anonymous.4open.science/r/illcond-FBB3/README.md}{Code: https://github.com/ChethanKodase/illcond}
\end{abstract}


\maketitle

\fancyhead{}
\fancyfoot{}
\renewcommand{\headrulewidth}{0pt}

\section{Introduction}
\label{sec:intro}
Autoencoders (AEs) \cite{Goodfellow-et-al-2016} are deployed in high-stakes applications, including image compression, reconstruction, denoising, anomaly detection, and generative modeling. By learning to approximately invert an encoder through a decoder, AEs inherently pose an inverse problem and exhibit structural non-invertibility due to dimensionality reduction.
Despite this, adversarial vulnerabilities of AEs remain less studied than those of classification models~\cite{szegedy2013intriguing}.
Rigorous robustness evaluation requires strong and well-understood white-box attack methodologies, which are a standard tool for robustness evaluation~\cite{akhtar2021ieee_survey}, as they are essential for identifying model vulnerabilities and designing effective defense ~\cite{van2024adversarial,szegedy2013intriguing}.
AEs are a prominent class of encoder–decoder architectures, motivating investigation of similar optimization challenges in broader encoder–decoder systems.

Existing white-box attacks for AEs~\cite{carlini2017towards,gondim2018adversarial,cemgil2020adversarially} optimize norm-bounded perturbations to maximize reconstruction damage. However, encoder–decoder architectures often exhibit 
ill-conditioning due to dimensionality reduction and approximate inversion during gradient propagation. As a result, adversarial optimization may yield suboptimal perturbations, potentially overstating robustness.


\noindent\textbf{Ill-Conditioning as a Source of Adversarial Vulnerability}. 
Ill-conditioning in neural networks can arise from highly imbalanced singular value spectra of the Jacobians of nonlinear network mappings, resulting in large condition numbers, as well as from structurally rank-deficient transformations induced by dimensionality reduction. Both effects are common in encoder–decoder architectures and can increase sensitivity to input perturbations. In practice, conditioning behavior is often approximated through the singular value spectra of intermediate layer weight matrices, which strongly influence gradient propagation \cite{sedghi2018singular,yoshida2017spectral,pennington2017resurrecting}. Prior studies~\cite{sinha2018neural,wei2022leccv_condition_no} have shown that ill-conditioned networks can amplify small perturbations and increase adversarial vulnerability. Research on instabilities in deep inverse problems~\cite{antun2021deep,antun2020instabilities} and the accuracy--stability trade-off~\cite{gottschling2025troublesome} further demonstrates that deep learning pipelines, including AEs, can exhibit severe ill-conditioning and increased sensitivity to perturbations. At the same time, near-zero singular values in ill-conditioned networks can suppress adversarial gradient propagation, degrading attack optimization.

\noindent\textbf{Why Small Singular Values Matter.}
Prior work on improving robustness has mainly focused on controlling large singular values through Lipschitz regularization and spectral normalization~\cite{barrett2022certifiably,fazlyab2019efficient}. In particular, several regularization methods~\cite{yoshida2017spectral,miyato2018spectral,cisse2017parseval,gulrajani2017improved} 
explicitly constrain spectral norms, i.e., the largest singular values of weight matrices~\cite{virmaux2018lipschitz}, to limit perturbation amplification and improve stability.
In addition, implicit self-regularization effects arising from optimization dynamics~\cite{martin2021implicit} have been observed to promote similar spectral control during training. 
In contrast, considerably less attention has been paid to near-zero singular values, which can induce rank-deficiency and instability~\cite{kirsch2011introduction}, suppress gradient propagation, and impair adversarial optimization, potentially preventing attacks from finding worst-case perturbations.

\noindent\textbf{Ill-Conditioning Behind the Illusion of Adversarial Robustness.}
Prior work on dynamical isometry \cite{pennington2017resurrecting} suggests that near-zero singular values break dynamical isometry, leading to ill conditioned Jacobians causing poor gradient flow and slow or stalled optimization. Similar effects were observed in quantized classification models during attack optimization~\cite{gupta2022improved}, creating the illusion of adversarial robustness by simply impeding adversarial optimization rather than improving true robustness.
However, the impact of ill-conditioning on gradient propagation during white-box adversarial optimization against encoder–decoder architectures remains largely unexplored. Our contributions are summarized as follows:\\
(i) We identify a previously unexplored failure mode in AE adversarial optimization, where near-zero singular values suppress adversarial gradients and yield suboptimal attacks.\\
(ii) We propose \ours{}, a technique that is designed to mitigate gradient degradation during adversarial optimization in ill-conditioned encoder--decoder architectures.
\\
(iii) We show that \ours{} consistently improves attack effectiveness and exposes vulnerabilities hidden by optimization limitations.\\
(iv) We provide preliminary evidence that modern multimodal encoder–decoder architectures may exhibit similar vulnerabilities.

Paper organization: Section~\ref{sec:related_work} reviews related work. Section~\ref{sec:preliminaries} introduces preliminaries and attack formulations.
We first introduce LGR in Section~\ref{sec:lgr} for simplified decoder-side ill-conditioning settings, before presenting \ours{} in Section~\ref{sec:ALMA} for general ill-conditioned encoder–decoder architectures.
Section~\ref{sec:experiments} reports experimental results and ablation studies, and Section~\ref{sec:conclusions} concludes the paper.


\section{Related Work}
\label{sec:related_work}
Related work spans four main areas: AE architectures, white-box adversarial attacks on autoencoders, adversarial attacks on broader encoder–decoder architectures, and optimization challenges arising from ill-conditioned networks.

\noindent\textbf{Architectures.}
Modern AEs extend the vanilla AE~\cite{hinton2006reducing} through structural and regularization enhancements. $\beta$-VAE~\cite{betaVAEoriginal} and TC-VAE~\cite{tcVAEOriginal} encourage disentangled latent representations via KL regularization and total correlation penalties, respectively. NVAE~\cite{nVAEoriginal} introduces hierarchical latent variables for multi-scale representations, while DiffAE~\cite{diffAEoriginal} integrates diffusion-based decoding to improve fidelity and robustness. Masked autoencoders (MAE)~\cite{he2022masked} employ self-supervised reconstruction via random masking without explicit latent regularization. While discrete-latent models exist~\cite{van2017neural}, we focus on state-of-the-art AEs with continuous latent spaces. 
To evaluate the broader applicability of \ours{} beyond classical AEs, we additionally consider large-scale vision–language encoder–decoder models, including Gemma~3~\cite{team2025gemma} and Qwen~2.5~\cite{hui2024qwen2}.

\noindent\textbf{White-box Attacks on Autoencoders.}
White-box adversarial attacks optimize norm-bounded perturbations with full access to model gradients and may be targeted or untargeted, we focus on the latter. Unlike targeted attacks such as C\&W~\cite{carlini2017towards}, which induce misclassification, untargeted attacks on AEs maximize latent or output distortion under a fixed perturbation budget using metrics such as L2 distance~\cite{gondim2018adversarial} or Wasserstein distance~\cite{cemgil2020adversarially}. Cosine similarity is commonly used in attacks on diffusion-based models~\cite{radford2021learning,zhuang2023pilot,zeng2024advi2i}. Symmetric KL divergence~\cite{camuto2021towards} has also been explored for probabilistic latent outputs, but is inapplicable to non-probabilistic intermediate and final AE representations and is therefore excluded. 
While classical white-box attacks probe intrinsic model robustness, adaptive attacks~\cite{tramer2020adaptive} evaluate defenses (e.g., Hamiltonian Monte Carlo-based defense~\cite{kuzina2022alleviating}) with full knowledge of the defense. 

\noindent\textbf{White-box Attacks on Encoder-Decoder Architectures.} Adversarial attacks on multimodal encoder–decoder architectures mainly focus on disrupting intermediate representations or maximizing prediction uncertainty. Dispersion Reduction Attack (DRA) \cite{lu2020enhancing},  Feature disruptive attack (FDA) \cite{ganeshan2019fda}, and Self Supervised Perturbation Attack (SSPA) \cite{naseer2020self} evaluate multimodal model robustness by perturbing inputs to distort latent feature spaces.  Blockwise Similarity Attack (BSA) \cite{yin2023vlattack} further extends this idea to vision-language models through block-wise feature disruption across image and transformer encoders. More recently, Entropy Guided Attacks (EGA) \cite{he2025few} showed that high-entropy tokens can act as critical failure points in vision-language models, leading to adversarial outputs. We use these methods as 
baselines to explore the applicability of \ours{} to broader encoder-decoder architectures.

\noindent\textbf{Vanishing Gradients from Ill-Conditioned Weights.}
Dynamical isometry via orthogonal initialization improves optimization stability in deep networks, highlighting the adverse impact of vanishing gradients during adversarial optimization~\cite{pennington2017resurrecting}. While gradient attenuation associated with near-zero singular values has been studied in quantized classification networks~\cite{gupta2022improved}, the proposed remedies rely on softmax outputs and are not applicable to AEs. Gradient vanishing has also been identified as a form of \emph{gradient obfuscation}, exploited by some defenses to hinder attack optimization~\cite{athalye2018obfuscated}. 
Similarly, ill-conditioned encoder–decoder architectures may hinder adversarial optimization through suppressed gradient propagation.
%
While prior work has explored hidden-layer perturbations~\cite{khrulkov2018art,tsymboi2023layerwise} and training-time spectral regularization~\cite{yoshida2017spectral,sedghi2018singular}, it does not address optimization-induced gradient degradation that can mask vulnerabilities during adversarial optimization in encoder–decoder architectures, which is our focus.

\section{Preliminaries }
\label{sec:preliminaries}
\noindent \textbf{Notation.}
Let each input sample be represented as \(x\in\mathbb{R}^{d}\).
An auto\-encoder $\mathcal{Y}:\mathbb{R}^d\rightarrow\mathbb{R}^d$ consists of an encoder $\phi$, which maps the input to a latent space $\mathbb{R}^n$, and a decoder $\psi$, which maps back to the original space $\mathbb{R}^d$ (typically, \(n < d\)):
\[
\mathcal{Y}(x)\;=\;\psi\circ\!\phi(x),\qquad
\phi:\mathbb{R}^{d}\!\to\!\mathbb{R}^{n},\;
\psi:\mathbb{R}^{n}\!\to\!\mathbb{R}^{d}
\]
Generally, for a function $f:\mathbb{R}^m\rightarrow\mathbb{R}^l$ and a vector $z\in\mathbb{R}^m$, we denote by $J^f_z\in\mathbb{R}^{l\times m}$ the Jacobian matrix of $f$ at $z$. We assume that all partial derivatives, and hence the Jacobian, exist throughout.

The $L_p$-ball around $x$ of radius $c$ is denoted by: $B^{p}_{c}(x)=\{x'\in\mathbb{R}^{d}\mid\lVert x'-x\rVert_{p}\le c\}$ where $c$ is the attack budget. 
Finally, a distortion measure $\Delta:\mathbb{R}^d\times\mathbb{R}^d\rightarrow\mathbb{R}$ is used (with $d$ replaced by $n$ for latent space distortions). 

\noindent \textbf{Multidimensional Chain Rule.}
The chain rule for higher dimensional maps states that the Jacobian matrix of a composed function $h=g\circ f$ is given by the matrix product of the Jacobians of $g$ and $f$. 
In particular, for an AE, the Jacobians of the encoder, and the decoder at input $x$ are denoted by $J^\phi_x$ and $J^\psi_{\phi(x)}$, respectively, yielding: $J^{\mathcal{Y}}_x=J^\psi_{\phi(x)} J^\phi_x$.

\noindent \textbf{Multidimensional Taylor Approximation.}
Assuming suitable smoothness, function $f:\mathbb{R}^m\rightarrow\mathbb{R}^l$ can be locally approximated at $z_a$ by 
\[
f(z_a) \simeq f(z)+J^f_z(z_a-z)+\mathcal{O}(||z_a-z||^2)
\]
where $\mathcal{O}(||z_a-z||^2)$ is the  remainder term, which decays quadratically. The approximation $f(z_a) \simeq f(z)+J^f_z(z_a-z)$ is commonly referred to as \emph{first-order approximation}.

\subsection*{Adversarial Attack Objectives}
Unlike targeted attacks such as C\&W~\cite{carlini2017towards}, which jointly optimize for misclassification and minimal perturbation magnitude, untargeted attacks on AEs are typically formulated as maximizing either output-space or latent-space distortion under a fixed perturbation budget~\cite{Tabacof2016AdversarialVAE,gondim2018adversarial,camuto2021towards}.

\noindent \textbf{Output–space Maximization (\textbf{OA})}
\emph{Output-space attacks} seek an adversarial input $x_a^* \in {B^{p}_{c}(x)}$ that maximizes reconstruction distortion between the original and adversarial inputs:
\begin{equation}\label{eq.out_goal}
  x_a^{\star}=\arg\max\limits_{x_a\in B^{p}_{c}(x)}
              \Delta\bigl(\mathcal{Y}(x_a),\mathcal{Y}(x)\bigr).
\end{equation}
Optimization: At iteration $t$, the adversarial input  $x_a^t$ is updated using gradient \emph{ascent} with step size \(\eta\), followed by projection onto the ball ${B^{p}_{c}(x)}$. 
This approach serves as a baseline method. 

\noindent \textbf{Latent–space Maximization (\textbf{LA})}
\emph{Latent-space attacks}~\cite{Tabacof2016AdversarialVAE} seek an adversarial input $x_a^* \in {B^{p}_{c}(x)}$ that maximizes distance between encoded latent representations:
\begin{equation}\label{eq.lat_goal}
  x_a^{\star}=\arg\max\limits_{x_a\in B^{p}_{c}(x)}
              \Delta\bigl(\phi(x_a),\phi(x)\bigr),
\end{equation}
Optimization: Similar to OA attacks, but gradients are computed using latent-space distortion rather than output-space distortion.
This serves as a second baseline method.


Existing works employ different distance metrics for untargeted attacks depending on the model architecture and training setting, including \(L_2\) distance~\cite{gondim2018adversarial}, Wasserstein distance~\cite{cemgil2020adversarially}, and cosine similarity in diffusion-based models~\cite{radford2021learning,zhuang2023pilot,zeng2024advi2i}. 

\subsection*{Vanishing Gradients and Ill-conditioning}

\noindent\textbf{Condition number.} For a matrix \(W\in \mathbb{R}^{m\times n}\), let
\(\sigma_{1}(W)\) and \(\sigma_{r}(W)\) denote its largest and smallest
non-zero singular values. Let  $r(W)\leq\min\{m,n\}$ denote its rank. The condition number of $W$ is defined as: \(\kappa(W)=\infty\) if $r(W)<n$, and 
\(
\kappa(W)=\sigma_{1}(W)/\sigma_{r}(W)\) otherwise.

A matrix is called \emph{ill-conditioned} when \(\kappa\gg1\). 
Large condition numbers indicate high sensitivity to perturbations along certain directions.
Neural network sensitivity has been linked to condition numbers of network parameters~\cite{sinha2018neural}. 

\noindent\textbf{Ill-conditioning in AEs}
Since we assume a smaller latent dimension $n<d$, the encoder Jacobian is structurally rank-deficient and therefore \emph{structurally ill-conditioned}, implying the existence of directions with reduced local sensitivity. Beyond the unavoidable ill-conditioning caused by  dimensionality reduction,  training can further amplify ill-conditioning in model layers (\emph{learned ill-conditioning}). 
In particular, training may drive singular values toward highly uneven magnitudes, with some becoming very large and others approaching zero.

Most adversarial attack objectives for AEs tend to implicitly exploit directions associated with large singular values, since under first-order approximation these directions maximize local amplification of norm-bounded perturbations and are closely related to local Lipschitz sensitivity~\cite{kuzina2021diagnosing,abuduweili2024estimating}. In contrast, small non-zero singular values may attenuate adversarial gradients and hinder optimization, motivating \ours{}.

\color{black}
\section{Latent Gradient Restoration under Decoder Ill-Conditioning}\label{sec:lgr}

There exists prior work on AE adversarial attacks optimizing only output-space divergence~\cite{gondim2018adversarial} (Eq.~\ref{eq.out_goal}), as well as attacks optimizing only latent-space divergence~\cite{kuzina2022alleviating}. Recent studies~\cite{barrett2022certifiably} show that certifiable robustness of AEs depends jointly on factors including the Lipschitz properties of the \emph{encoder} and \emph{decoder}~\cite{hager1979lipschitz}, as well as the sensitivity of latent representations to input perturbations. This suggests that adversarial vulnerability in encoder--decoder architectures depends not only on reconstruction-space behavior, but also on the stability of latent representations and the propagation of gradients through both components. However, existing adversarial objectives do not account for optimization difficulties induced by ill-conditioned layers. 

We first consider a simplified setting in which the decoder contains ill-conditioned intermediate layers while the encoder remains well-conditioned. In this setting, decoder-side ill-conditioning can attenuate adversarial gradients during backpropagation. To address this issue, we introduce \emph{Latent Gradient Restoration} (\oursMini), which couples latent-space and reconstruction-space distortions such that latent-space gradients compensate for degraded reconstruction-space gradients, while reconstruction-space distortion continues to guide optimization by scaling the latent-space gradient.

We define the following maximum-damage criterion:
\begin{equation}\label{eq.LMA}
    \begin{split}
         x_a^* = \arg\max\limits_{ x_a\in B^p_c(x)}\mathcal{L}(x_a).
    \end{split}
\end{equation}

where $\mathcal{L}(x_a)$ is defined as:
\begin{equation}\label{eq.LMA_product}
    \begin{split}
         \mathcal{L}(x_a)= \Delta\bigl(\phi(x_a),\phi(x)\bigr) \cdot
         \Delta\bigl(\psi \circ \phi(x_a),\psi \circ \phi(x)\bigr).
    \end{split}
\end{equation}

The product structure yields the following gradient:
\begin{equation}\label{eq.gradRestore}
\begin{aligned}
\nabla_{x_a} \mathcal{L} =\;&
\Delta(\phi(x_a), \phi(x)) \cdot \nabla_{x_a} \Delta\bigl(\psi \circ \phi(x_a),\psi \circ \phi(x)\bigr) \\
&+ \Delta\bigl(\psi \circ \phi(x_a),\psi \circ \phi(x)\bigr) \cdot \nabla_{x_a} \Delta(\phi(x_a), \phi(x)).
\end{aligned}
\end{equation}


\paragraph{Gradient Degradation Mitigation Mechanism}
When the decoder output gradient $\nabla_{x_a}\Delta\bigl(\psi \circ \phi(x_a),\psi \circ \phi(x)\bigr) \to 0$
due to ill-conditioned decoder layers, its direct propagation becomes weak. However, since the encoder remains well-conditioned, the latent-space gradient
$\nabla_{x_a}\Delta(\phi(x_a), \phi(x))$
remains non-negligible. Consequently, the term
\[
\Delta\bigl(\psi \circ \phi(x_a),\psi \circ \phi(x)\bigr)
\cdot
\nabla_{x_a}\Delta(\phi(x_a), \phi(x))
\]
continues to provide a non-vanishing ascent direction, while the output-space distortion
$\Delta\bigl(\psi \circ \phi(x_a),\psi \circ \phi(x)\bigr)$
still influences optimization by scaling non--vanishing latent--space gradient \\$\nabla_{x_a}\Delta(\phi(x_a), \phi(x))$, analogous to gradient modulation strategies studied in auxiliary-loss optimization~\cite{du2018adapting}. Thus, gradient degradation caused by the ill-conditioned decoder is mitigated through latent-space gradients originating from the well-conditioned encoder, resulting in non-trivial adversarial updates.

\begin{figure}[t]
    \centering
    \begin{minipage}[b]{0.48\textwidth}
        \centering
        \includegraphics[width=\linewidth]{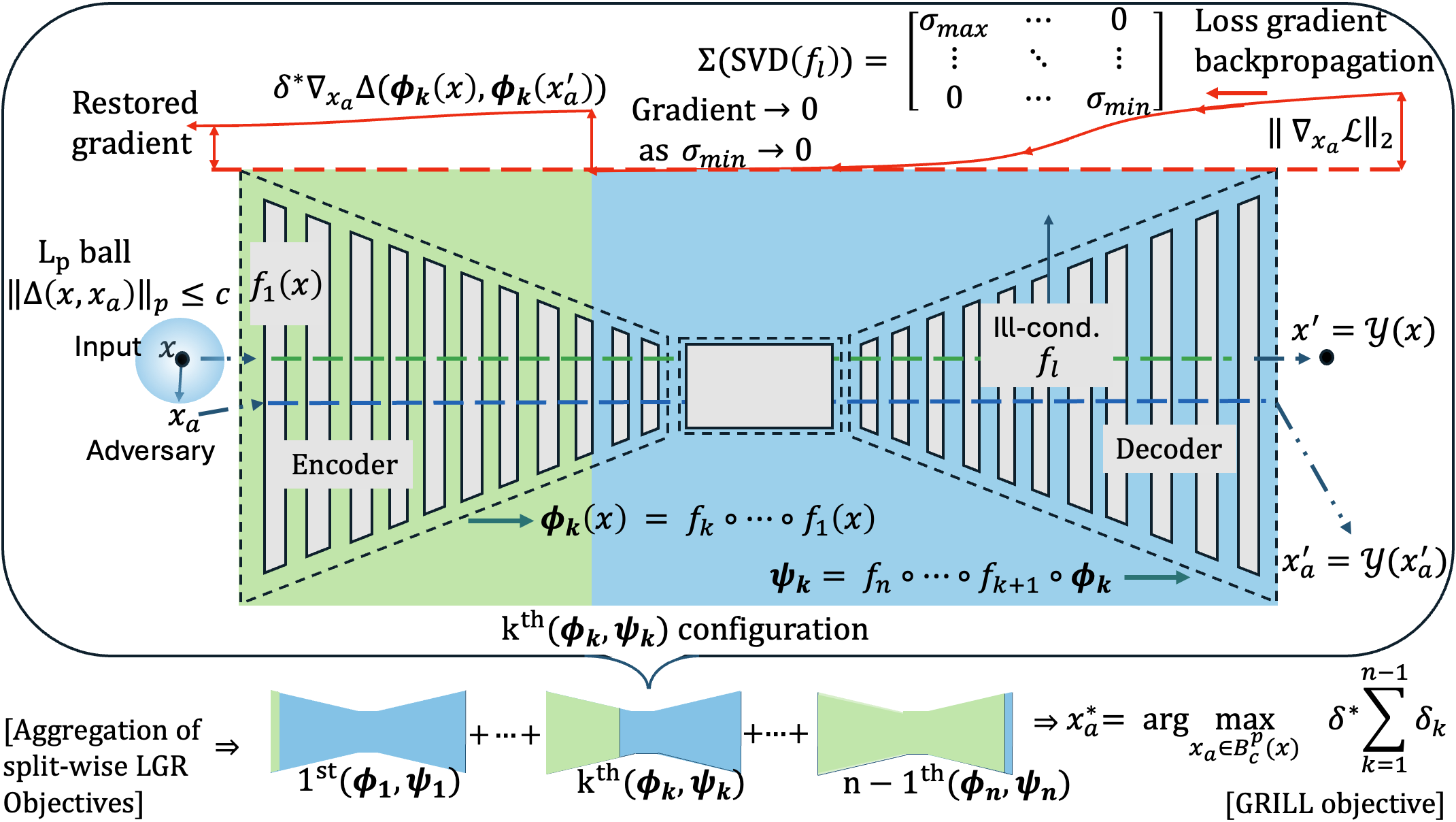}
    \end{minipage}\hfill
    \caption{\ours{} framework for mitigating gradient degradation in ill-conditioned encoder-decoder architectures.}
    \label{fig:GRILLarchitecture}  
\end{figure}

Importantly, using a product rather than a sum introduces multiplicative coupling between latent-space and reconstruction-space distortions, allowing each distortion magnitude to scale the gradient contribution of the other. This explicit cross-weighting can improve optimization dynamics relative to naive loss summation (shown empirically in Section~\ref{sec:expAblationEfficiency}).
Figure~\ref{fig:GRILLarchitecture} illustrates the GRILL framework and the role of split-wise latent gradient restoration.

\section{\oursName}\label{sec:ALMA}
\oursMini{} (Section~\ref{sec:lgr}) considers a simplified setting where gradient degradation originates from ill-conditioned decoder layers while the encoder remains well-conditioned. However, in practical encoder--decoder architectures, ill-conditioning can occur anywhere in the network, including intermediate encoder layers, limiting the applicability of \oursMini{}. To address this issue, we introduce \ours{} (\oursName), which considers the AE $\mathcal{Y}$ as a composition of $n$ layers \(\mathcal{Y} = f_n \circ \dots \circ f_1\) inducing \(n-1\) possible \emph{encoder-decoder} configurations (splits) indexed by \(k\in \{1,\dots,n-1\}\). Specifically, each ``split'' is given by:
\begin{itemize}
    \item \(\boldsymbol{\phi}_k = f_k \circ f_{k-1} \circ \dots \circ f_1\), \quad serving as an encoder up to layer \(k\),
    \item \(\boldsymbol{\psi}_k = f_n \circ f_{n-1} \circ \dots \circ f_{k+1},\) \quad serving as a decoder from layer \(k+1\) to \(n\).
\end{itemize}

The latent distortion after layer \(f_k\) can thus be expressed  in terms of $\boldsymbol\phi_k$ as 
$$
\delta_k = \Delta(\boldsymbol{\phi}_k(x_a), \boldsymbol{\phi}_k(x)).
$$
Since \(\boldsymbol{\psi}_k\) takes the representations produced by \(\boldsymbol{\phi}_k\) as input, the corresponding reconstruction-space distortion for the same split is
$$
\delta_k^* = \Delta\bigl(\boldsymbol\psi_k \circ\boldsymbol\phi_k(x_a), \boldsymbol\psi_k \circ \boldsymbol\phi_k(x)\bigr) .
$$
We therefore consider the $n-1$ splits and define the \ours{} loss as

\begin{equation}\label{eq.grill_final_before_common}
    \mathcal{L}_{\text{\ours{}}}(x_a)
    =
    \sum_{k=1}^{n-1}
    \delta_k \delta_k^*.
\end{equation}
Since the decoders of all splits reconstruct the same AE output, we consider $\delta_k^* = \delta^*=\Delta(\mathcal{Y}(x_a),
\mathcal{Y}(x)).$ The final optimization objective becomes
\begin{equation}
 \begin{split}
    \label{eq.grill_final}
    {x_a}^* =
    \arg\max\limits_{x_a\in{B}_c^p(x)}
    \delta^*
    \sum_{k=1}^{n-1}\delta_k.
\end{split}
\end{equation}
This objective is optimized using a gradient-based update scheme under an \(L_p\)-norm perturbation constraint. 

\subsection{How \ours{} Differs from Layer-wise Loss Summation}

A Layer-wise Loss Summation (LLS)-based attack typically aggregates intermediate distortions additively:
\begin{equation}
\mathcal{L}_{\mathrm{LLS}}(x_a)
=
\sum_{k=1}^{n-1}\delta_k(x_a),
\end{equation}
yielding
\begin{equation}
\nabla_{x_a}\mathcal{L}_{\mathrm{LLS}}
=
\sum_{k=1}^{n-1}\nabla_{x_a}\delta_k.
\end{equation}

Under severe ill-conditioning, gradients propagated through deeper transformations may become strongly attenuated, causing the corresponding terms \(\nabla_{x_a}\delta_k\) to approach zero. Consequently, additive aggregation cannot preserve the influence of reconstruction-space distortion once gradient degradation occurs. 
In contrast, GRILL introduces multiplicative coupling:
\begin{equation}
\mathcal{L}_{\mathrm{GRILL}}(x_a)
=
\delta^\ast(x_a)\sum_{k=1}^{n-1}\delta_k(x_a),
\end{equation}
with gradient
\begin{equation}
\nabla_{x_a}\mathcal{L}_{\mathrm{GRILL}}
=
\left(
\sum_{k=1}^{n-1}\delta_k
\right)
\nabla_{x_a}\delta^\ast
+
\delta^\ast
\sum_{k=1}^{n-1}\nabla_{x_a}\delta_k.
\end{equation}

Unlike additive aggregation, GRILL scales intermediate-layer gradients by the reconstruction distortion \(\delta^\ast\). As a result, reconstruction-space distortion continues to influence attack optimization by scaling non-vanishing gradients from less ill-conditioned representations, helping mitigate optimization stagnation under degraded gradient propagation.
Section~\ref{sec:expAblationEfficiency} empirically confirms that removing the reconstruction distortion term $\delta^*$ from Eq.\ref{eq.grill_final} reduces the formulation to simple layer-loss summation (LLS), which generates weaker attacks than \ours{}.

Algorithm~\ref{alg:grill_universal} summarizes the optimization procedure of \ours{} under an \(L_p\)-norm perturbation constraint.
\begin{algorithm}[t]
\caption{Universal adversarial attack with $L_\infty$ bound on perturbation using \ours{}}
\label{alg:grill_universal}
\textbf{Input}: Test set $\mathcal{D} = \{x_i\}_{i=1}^N$, trained AE $\mathcal{Y} = \psi \circ \phi = f_n \circ \dots \circ f_1$, for a configuration of  ($\boldsymbol\psi_k, \boldsymbol\phi_k$), \(\boldsymbol{\phi}_k = f_k \circ f_{k-1} \circ \dots \circ f_1\), \quad serves as an encoder up to layer \(k\),  \(\boldsymbol{\psi}_k = f_n \circ f_{n-1} \circ \dots \circ f_{k+1},\) \quad serves as a decoder from layer \(k+1\) to \(n\), step size $\eta$, number of steps till convergence $T$, batch size $B$ \\
\textbf{Parameter}: Perturbation budget $c$ is the $L_\infty$ bound \\
\textbf{Output}: Universal adversarial perturbation $\rho$
\begin{algorithmic}[1]
\STATE Initialize perturbation $\rho$ as a near-zero tensor with small random noise:
\STATE \hspace{1em} $\rho \sim \mathcal{U}(-\xi, \xi)$, where $\xi \ll c$
\FOR{$t = 1$ to $T$}
    \FOR{each batch $\{x_j\}_{j=1}^B$ in $\mathcal{D}$}
        \STATE Compute adversarial batch: $x_j^{\text{adv}} \leftarrow x_j + \rho$
        \STATE Compute \ours{} loss:
        \STATE \hspace{1em} $\delta_k \leftarrow \sum_{j=1}^{B} \Delta(\boldsymbol{\phi}_k(x_j^{\text{adv}}), \boldsymbol{\phi}_k(x_j^{\text{ }}))$
        \STATE \hspace{1em} $\delta^* \leftarrow \sum_{j=1}^{B}\Delta(\mathcal{Y}(x_j^{\text{adv}}), \mathcal{Y}(x_j^{\text{ }}))$
        \STATE \hspace{1em} $\mathcal{L}_{\text{adv}} \leftarrow \delta^* \cdot \sum_{k=1}^{n-1} \delta_k$
        \STATE Compute gradient: $g \leftarrow \nabla_{\rho} \mathcal{L}_{\text{adv}}$
        \STATE Update perturbation using Adam optimizer:
        \STATE \hspace{1em} $\rho \leftarrow \text{AdamStep}(\rho, \nabla_\rho \mathcal{L}_{\text{adv}})$
        \STATE Project onto $L_\infty$ ball:
        \STATE \hspace{1em} $\rho \leftarrow \mathrm{clip}(\rho, -c, c)$
    \ENDFOR
\ENDFOR
\STATE \textbf{return} $\rho$
\end{algorithmic}
\end{algorithm}
By aggregating \oursMini{} across all \emph{encoder-decoder} splits (line 9), \ours{} alleviates gradient attenuation caused by ill-conditioned encoder layers, since different splits can still contribute non-vanishing adversarial gradients as described in Section~\ref{sec:lgr} and shown in Figure~\ref{fig:GRILLarchitecture}.

\section{Experiments}
\label{sec:experiments}
\begin{figure}[t]
\centering

\begin{subfigure}[b]{0.40\linewidth}
  \includegraphics[width=\linewidth]{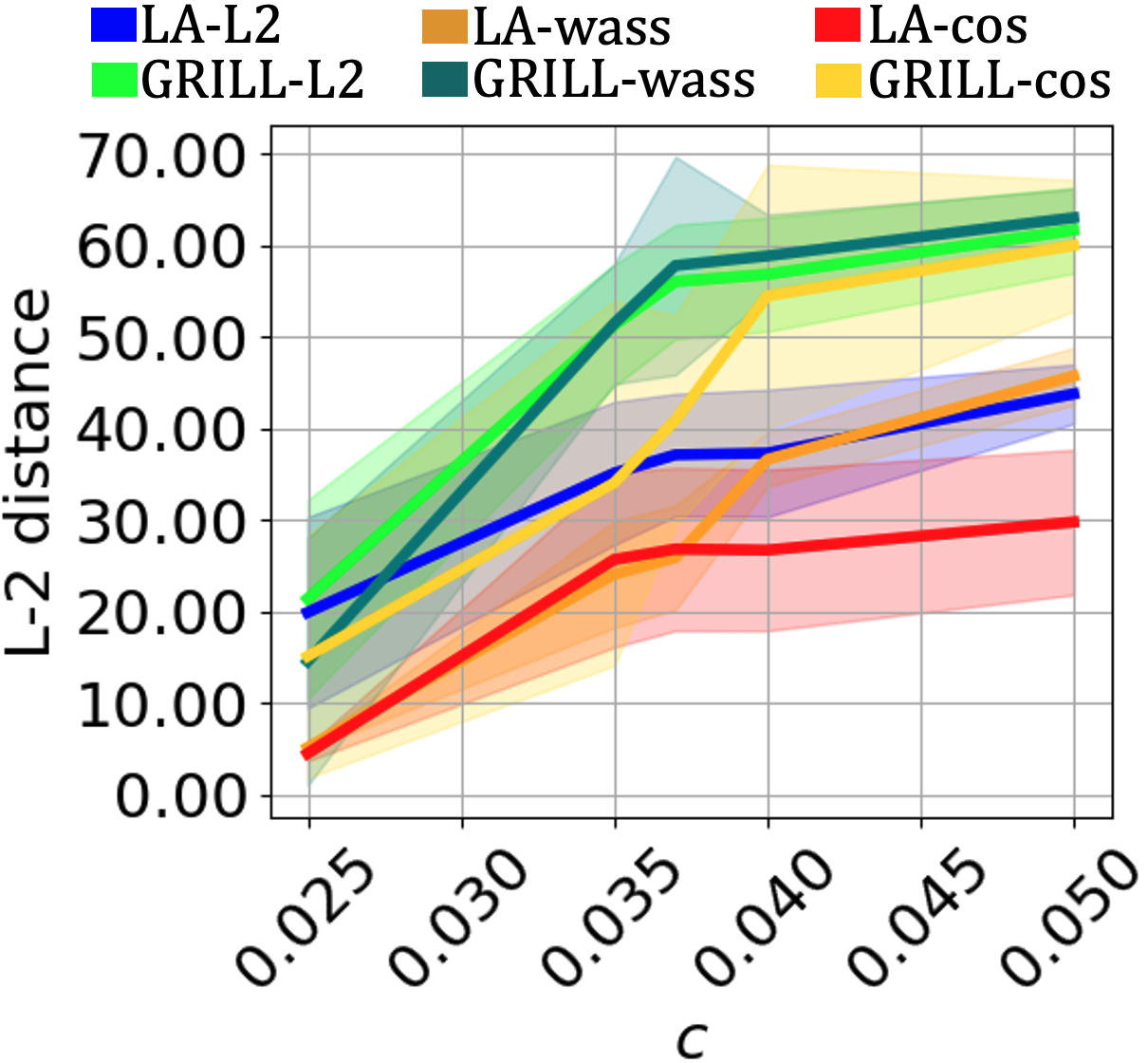}
  \subcaption{NVAE: OD vs. $c$}
  \label{fig:nvae_var}
\end{subfigure}\hfill
\begin{subfigure}[b]{0.53\linewidth}
  \includegraphics[width=\linewidth]{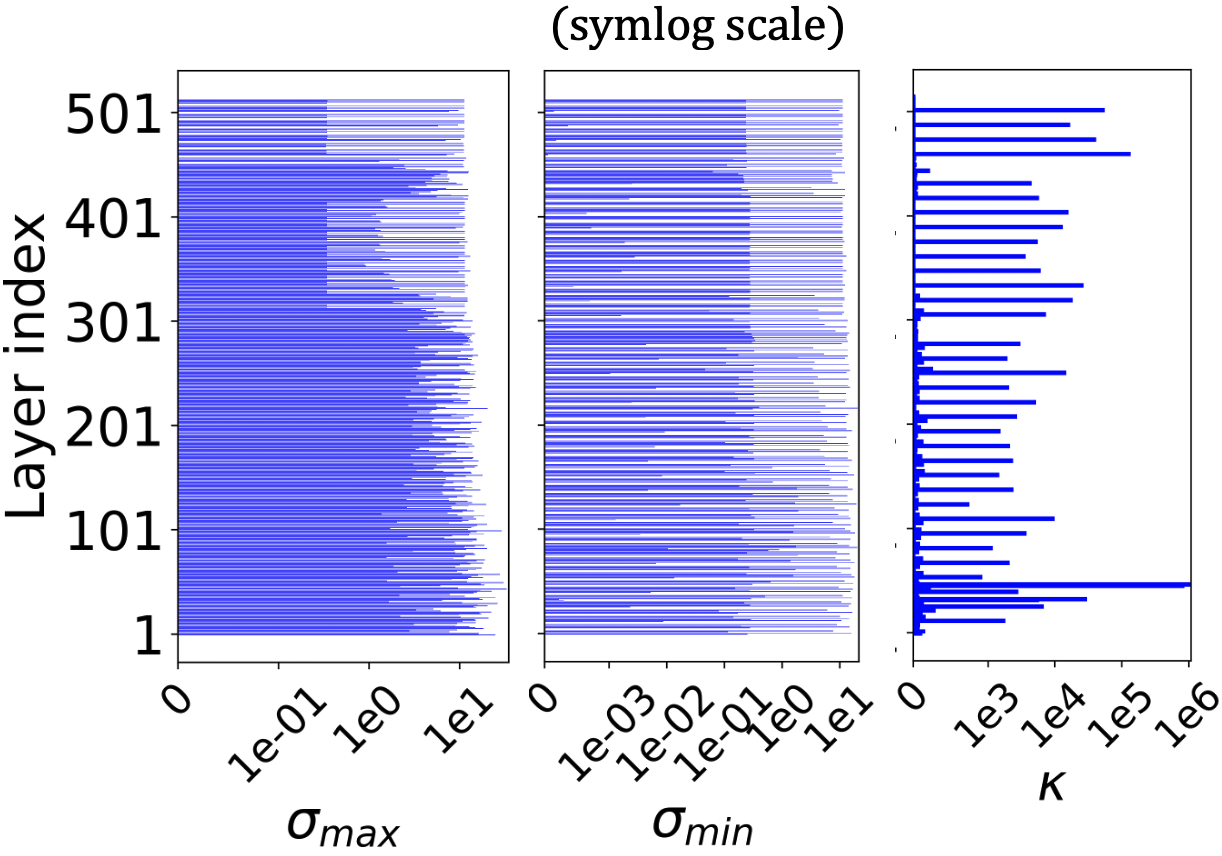}
  \subcaption{NVAE: $\sigma_{\max}$, $\sigma_{\min}$, $\kappa$}
  \label{fig:nvae_s_s_k}
\end{subfigure}

\vspace{0.6em}

\begin{subfigure}[b]{0.40\linewidth}
  \includegraphics[width=\linewidth]{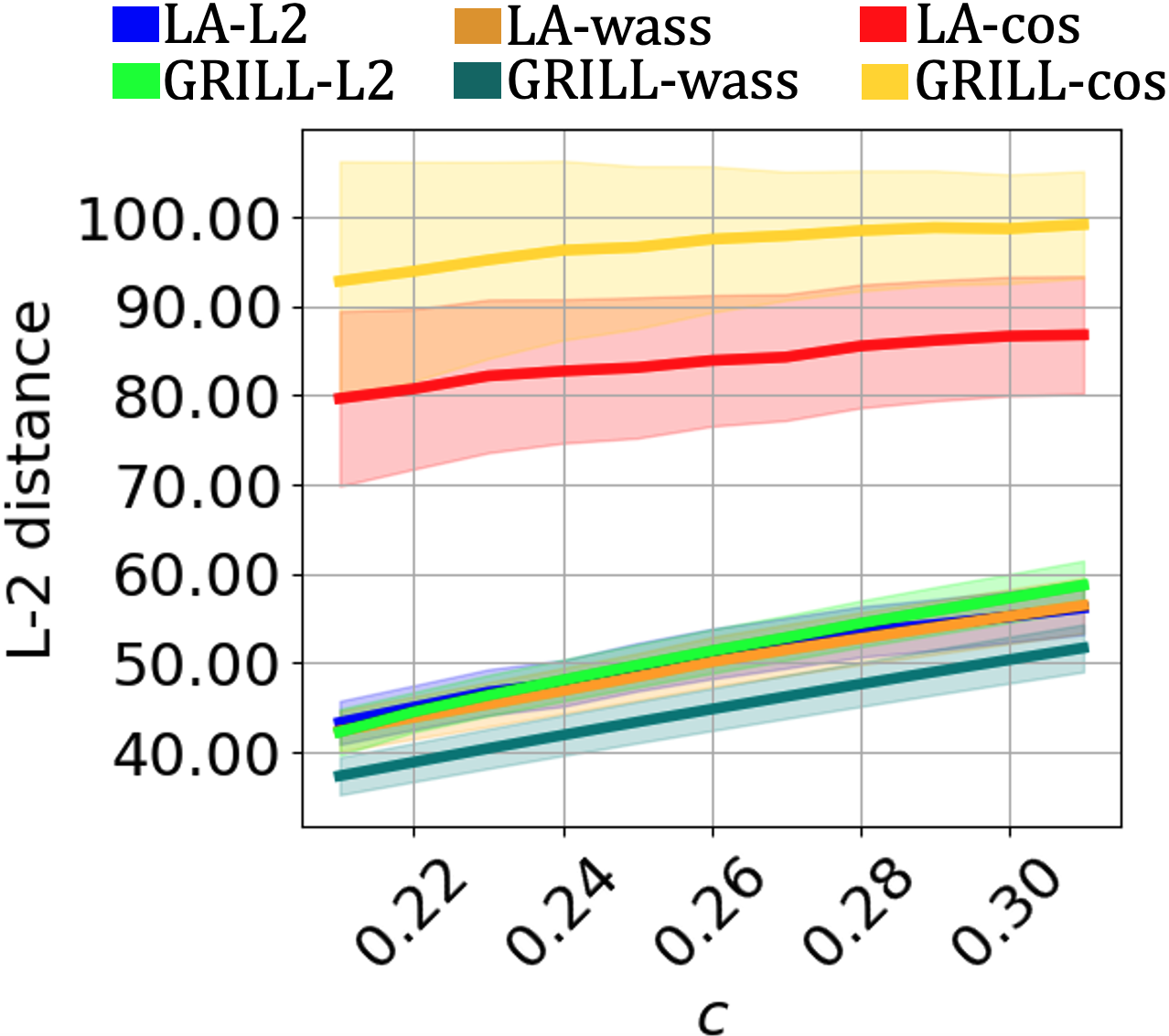}
  \subcaption{DiffAE: OD vs. $c$}
  \label{fig:diffae_var}
\end{subfigure}\hfill
\begin{subfigure}[b]{0.53\linewidth}
  \includegraphics[width=\linewidth]{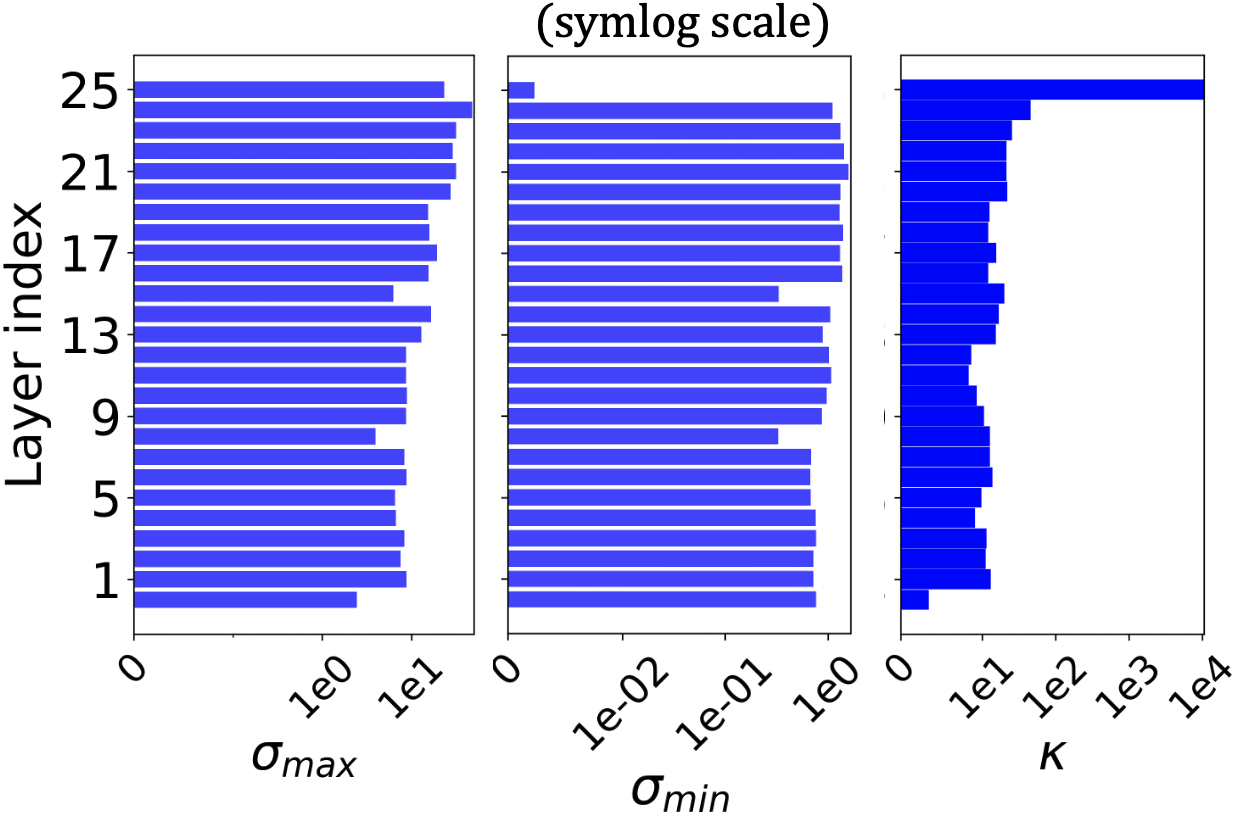}
  \subcaption{DiffAE: $\sigma_{\max}$, $\sigma_{\min}$, $\kappa$}
  \label{fig:diffae_s_s_k}
\end{subfigure}

\caption{
Universal attack performance in terms of output distortion (OD) (left) and corresponding layer-wise conditioning profiles (right) for highly ill-conditioned models.}
\label{fig:universal_high_ill}
\end{figure}

\begin{figure}[t]
\centering

\begin{subfigure}[b]{0.40\linewidth}
  \includegraphics[width=\linewidth]{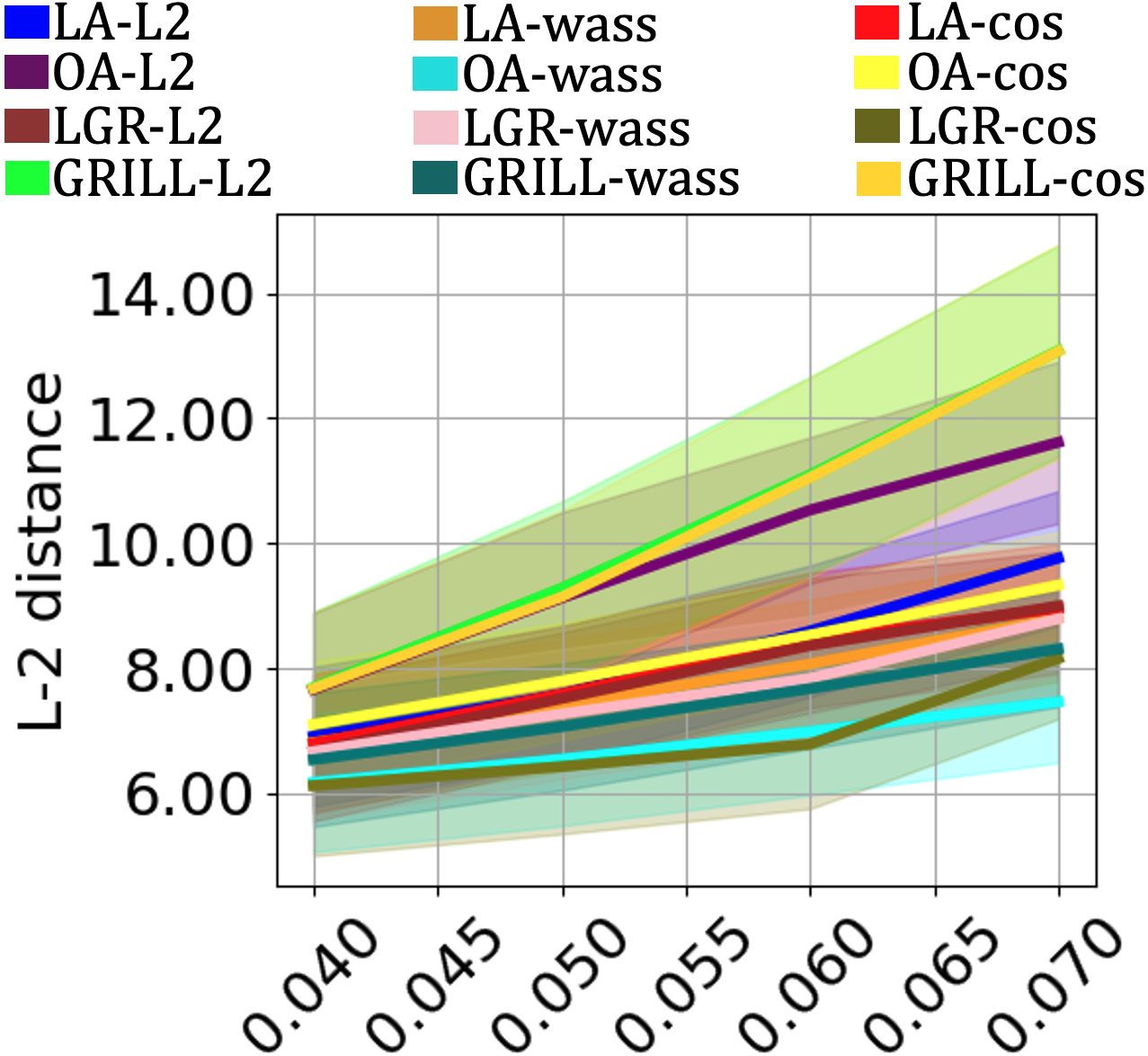}
  \subcaption{TC-VAE: OD vs. $c$}
  \label{fig:tc_vae_var}
\end{subfigure}\hfill
\begin{subfigure}[b]{0.55\linewidth}
  \includegraphics[width=\linewidth]{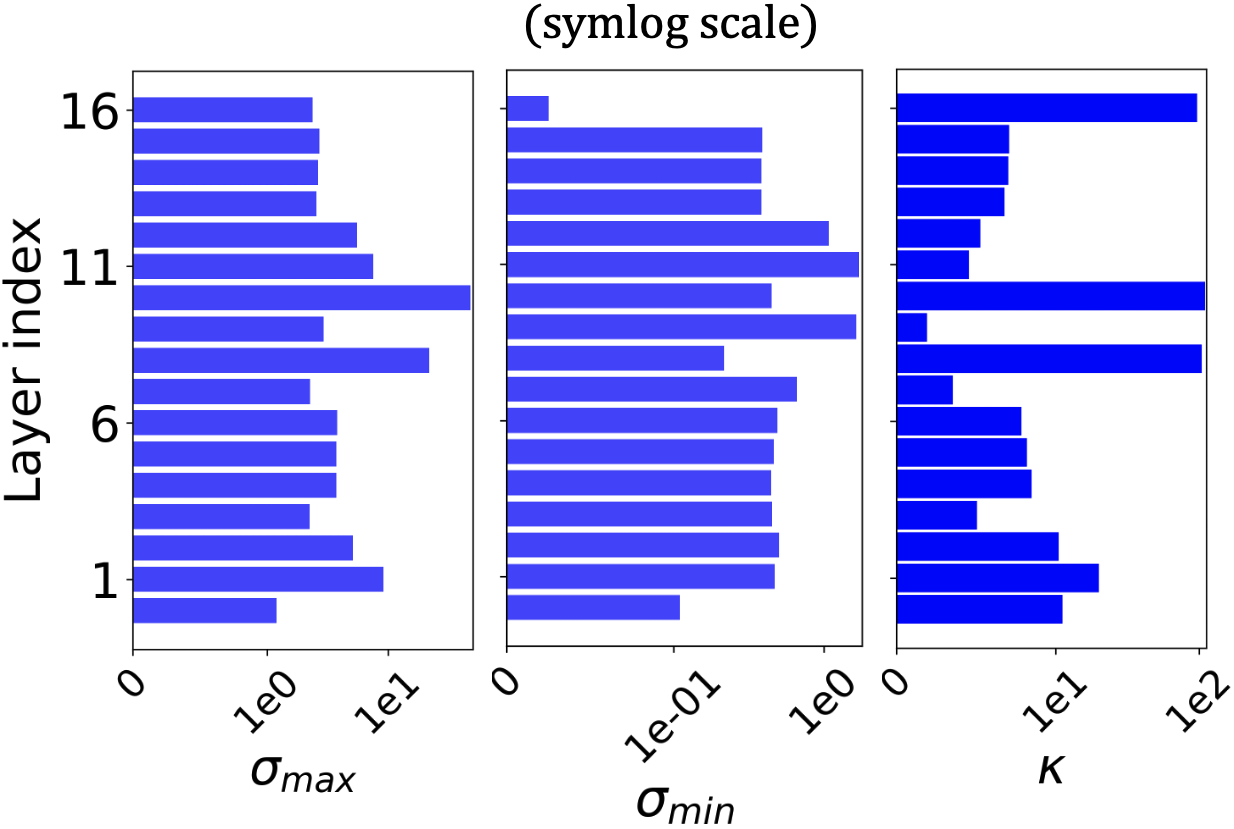}
  \subcaption{TC-VAE: $\sigma_{\max}$, $\sigma_{\min}$, $\kappa$}
  \label{fig:tc_vae_s_s_k}
\end{subfigure}

\vspace{0.6em}

\begin{subfigure}[b]{0.40\linewidth}
  \includegraphics[width=\linewidth]{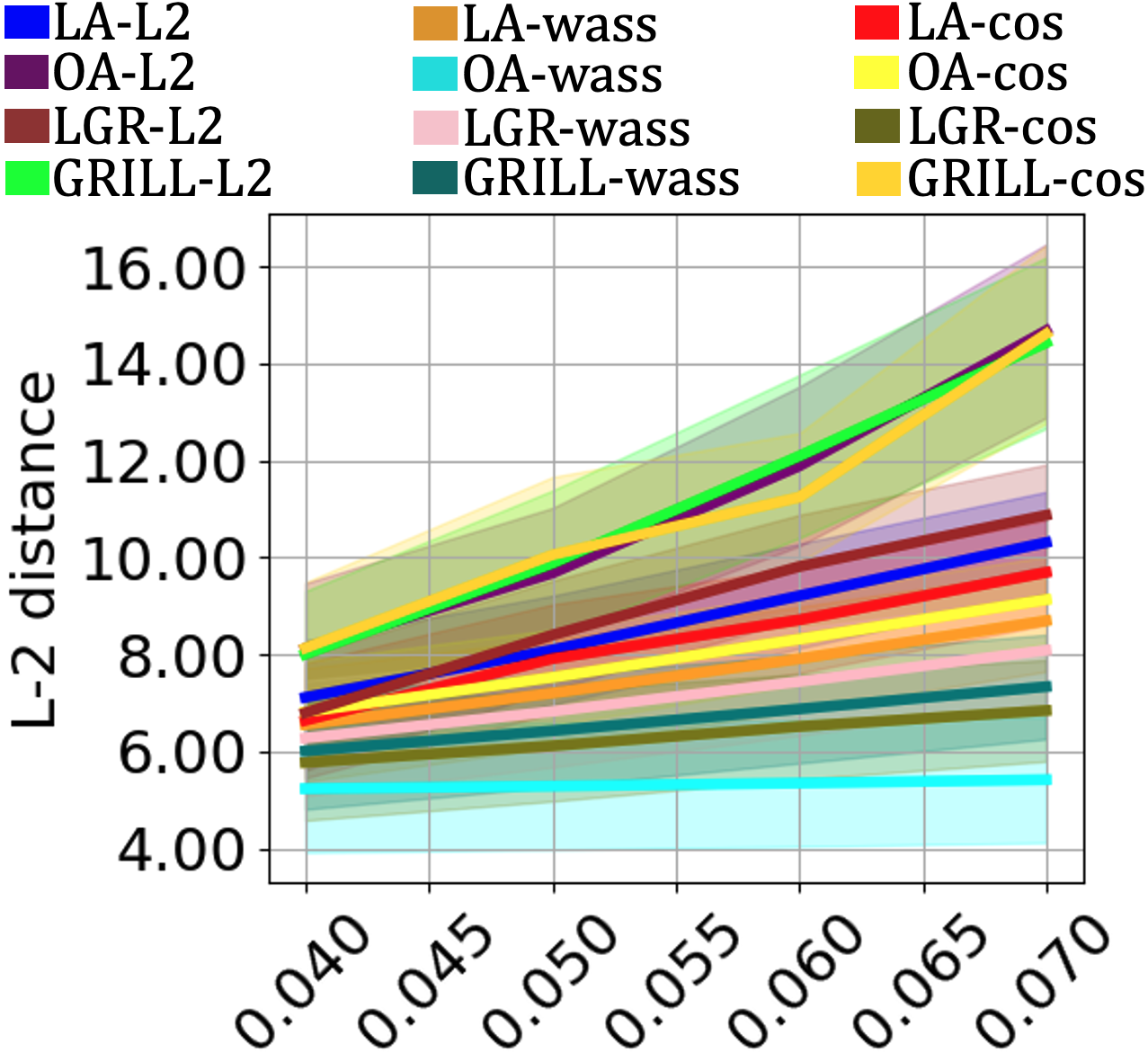}
  \subcaption{$\beta$-VAE: OD vs. $c$}
  \label{fig:beta_vae_var}
\end{subfigure}\hfill
\begin{subfigure}[b]{0.55\linewidth}
  \includegraphics[width=\linewidth]{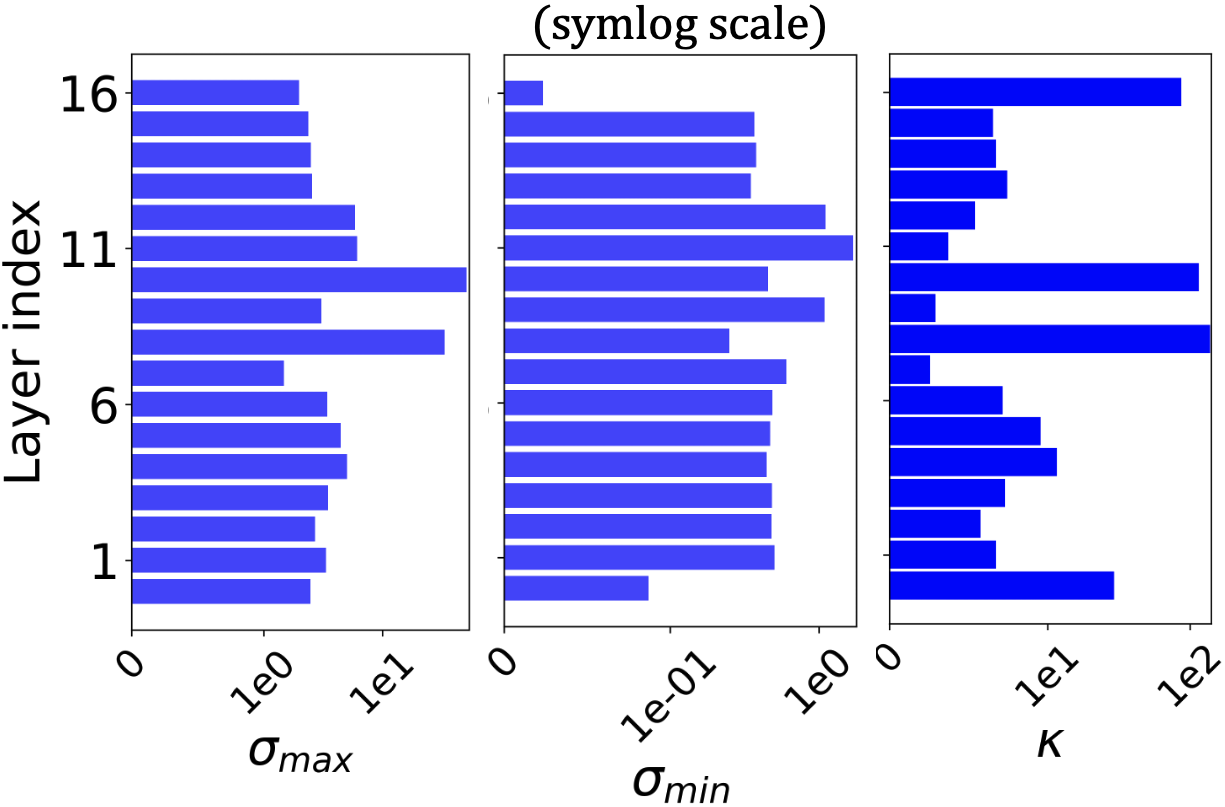}
  \subcaption{$\beta$-VAE: $\sigma_{\max}$, $\sigma_{\min}$, $\kappa$}
  \label{fig:beta_vae_s_s_k}
\end{subfigure}

\vspace{0.6em}

\begin{subfigure}[b]{0.40\linewidth}
  \includegraphics[width=\linewidth]{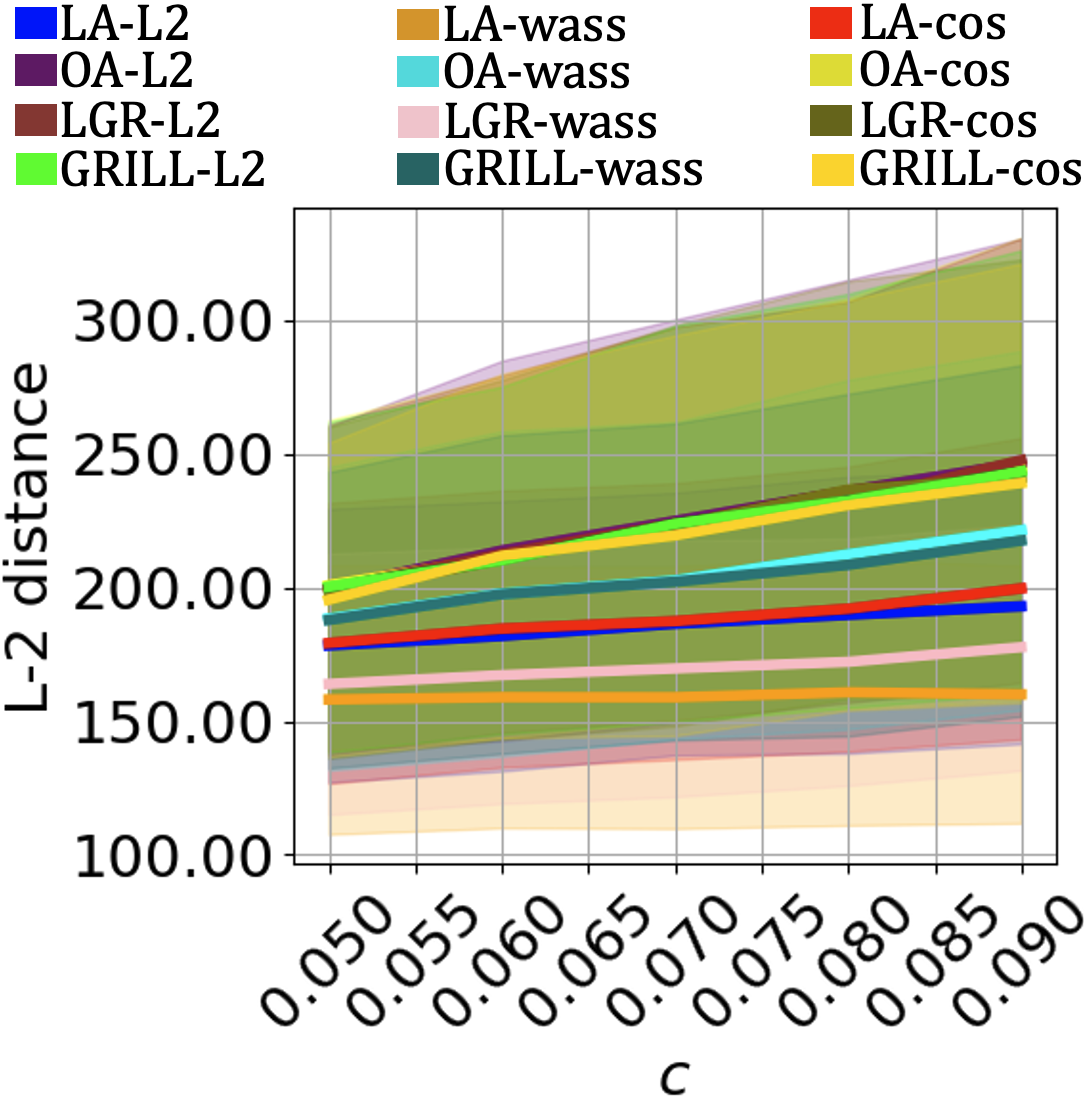}
  \subcaption{MAE: OD vs. $c$}
  \label{fig:mae_var}
\end{subfigure}\hfill
\begin{subfigure}[b]{0.55\linewidth}
  \includegraphics[width=\linewidth]{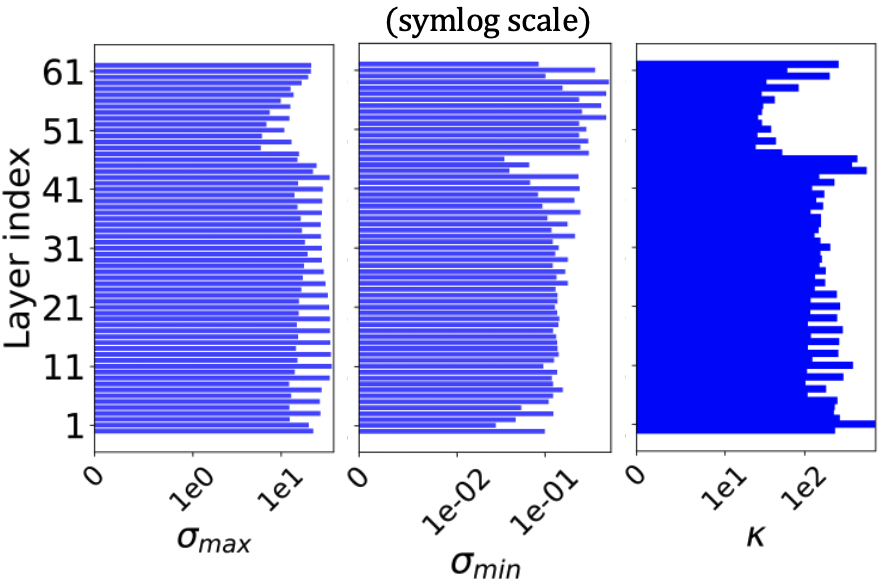}
  \subcaption{MAE: $\sigma_{\max}$, $\sigma_{\min}$, $\kappa$}
  \label{fig:mae_s_s_k}
\end{subfigure}

\caption{
Universal attack performance in terms of output distortion (OD) (left) and corresponding layer-wise conditioning profiles (right) for mildly ill-conditioned models.}
\label{fig:universal_mod_ill}
\end{figure}

Section~~\ref{sec:expSettings} presents the experimental setup and evaluation protocol. Section~\ref{sec:expConditioningBehaviour} analyzes the conditioning behavior of the evaluated models. Sections~\ref{sec:expUniversalAttacksAAEs} and~\ref{sec:expSampleSpecificAttackesAE} evaluate universal and sample-specific attacks on AEs, respectively, while Section~\ref{sec:expAttacksVLMs} studies attacks on VLMs. Finally, Section~\ref{sec:expAblationEfficiency} presents ablation studies, convergence and efficiency results.


\subsection{Experimental Setup}
\label{sec:expSettings}

We evaluate \ours{} across five AEs and two vision-language models (Gemma 3 (4.3B parameters) \cite{team2025gemma} and Qwen 2.5 (8.3B parameters) \cite{hui2024qwen2}) under multiple adversarial attack settings, perturbation budgets, and distance metrics (Section~\ref{sec:expSettings}). Table~\ref{tab:models_datasets_radii} summarizes the experimental setup. 
While introduced for AEs, \ours{} targets gradient propagation through deep nonlinear encoder–decoder transformations, motivating exploratory evaluation on modern VLMs~\cite{tan2019lxmert,li2022blip}. The datasets span both face-centric data (CelebA, FFHQ) and large-scale natural image data (ImageNet), aligned with the respective training or pretraining setups. Perturbation magnitudes were selected via grid search to balance attack effectiveness and  imperceptibility.

\begin{table}[t]
    \centering
    \small
    \setlength{\tabcolsep}{6pt}
    \begin{tabular}{lccc}
    \toprule
    \textbf{Model} & \textbf{Dataset} & \textbf{Training} & \textbf{$L_\infty$ Radius $c$} \\
    \midrule
    $\beta$-VAE   & CelebA\cite{liu2015faceattributes}   & From scratch & $[0.04, 0.07]$ \\
           \midrule
    TC-VAE        & CelebA   & From scratch & $[0.04, 0.07]$ \\
        \midrule
    NVAE\footnote{\url{https://github.com/NVlabs/NVAE}}          & CelebA   & Pretrained   & $[0.025, 0.05]$ \\
        \midrule
    DiffAE\footnote{\url{https://github.com/phizaz/diffae}}        & FFHQ\cite{karras2019style}     & Pretrained   & $[0.21, 0.30]$ \\
            \midrule
    MAE\footnote{\url{https://github.com/facebookresearch/mae}}           & ImageNet \cite{deng2009imagenet}  & Pretrained   & $[0.05, 0.09]$ \\
                \midrule    Gemma~3\footnote{\url{https://www.kaggle.com/models/google/gemma-3}}       & ImageNet & Pretrained & $[0.0005, 0.0009]$ \\
        \midrule
    Qwen~2.5\footnote{\url{https://www.kaggle.com/models/qwen-lm/qwen2.5}}      & ImageNet & Pretrained & $[0.001, 0.005]$ \\
    \bottomrule
    \end{tabular}
    \caption{Models, datasets, training setup, and perturbations.}
    \label{tab:models_datasets_radii}
\end{table}
\begin{figure}
    \centering
\includegraphics[width=1.0\linewidth]{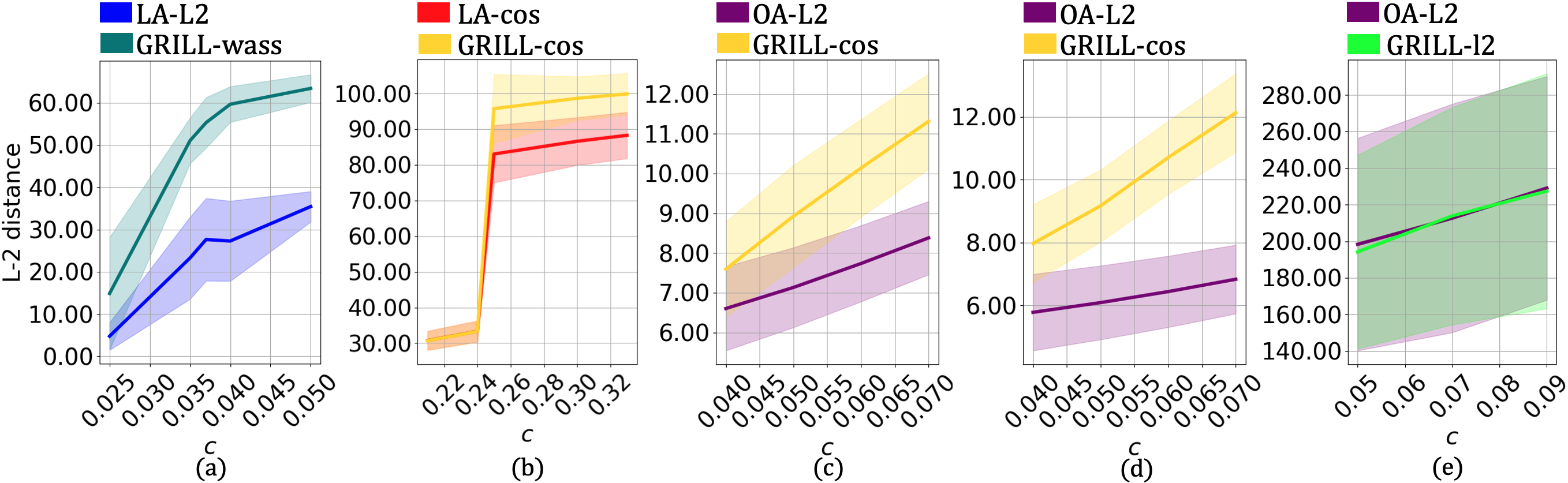}
\caption{Universal adaptive attack performance in terms of output distortion (OD) under varying perturbation radii $c$ for NVAE (a), DiffAE (b), TC-VAE (c), $\beta$-VAE (d), and MAE (e).}\label{fig:adaptive}
\end{figure}
\noindent \textbf{Attack Methods:} We unify the existing AE attack formulations in \cite{gondim2018adversarial,cemgil2020adversarially,radford2021learning,zhuang2023pilot,zeng2024advi2i} into the following AE attack baseline methods for comparison with \ours{}:\\
\noindent \textbf{- OA} (Output-space attack): maximize output-space distortion(Eq.~\ref{eq.out_goal}).\\
\noindent \textbf{- LA} (Latent-space attack): maximize latent-space distortion (Eq.~\ref{eq.lat_goal}); for NVAE, all hierarchical latent layers are attacked jointly~\cite{willetts2019improving}\\
\noindent \textbf{-\oursMiniName~(\oursMini):} improves adversarial gradient propagation in decoder-only ill-conditioned settings (Eq.~\ref{eq.LMA}).\\
\noindent \textbf{-\oursName~(\ours{}):} improves adversarial gradient propagation across intermediate encoder--decoder splits.(Eq.~\ref{eq.grill_final}).


\noindent\textbf{Distance Functions.} As discussed in Section~\ref{sec:related_work}, no single distance is universally effective, so we evaluate three common distance functions for adversarial loss: $L_2$~\cite{gondim2018adversarial}, Wasserstein (wass), and cosine similarity (cos)~\cite{radford2021learning,zhuang2023pilot,zeng2024advi2i}.
Attack configurations are defined by combining the attack objective (OA, LA, LGR, \ours{}) and distance function (L2, wass, cos).
For $\beta$-VAE, TC-VAE, and MAE, we evaluate all attack configurations. For NVAE and DiffAE, where decoding is computationally expensive, we limit evaluation to LA and \ours{}, both operating solely on the encoder. 

\noindent \textbf{VLM Attack Strategies.}
For VLMs, we apply \ours{} using intermediate vision and language representations within the model, and compare against representative multimodal attack baselines including \textbf{DRA}~\cite{lu2020enhancing}, \textbf{FDA}~\cite{ganeshan2019fda}, \textbf{SSPA}~\cite{naseer2020self}, \textbf{BSA}~\cite{yin2023vlattack}, and \textbf{EGA}~\cite{he2025few}.

\noindent\textbf{Evaluation measures.} Following standard practice~\cite{sakurada2014anomaly,willetts2019improving,our2024adversarial},  AE attacks are evaluated using \(L_2\) output distortion over 2,000 samples per attack configuration.
For VLMs, we report qualitative results for sample-specific attacks and additionally perform quantitative evaluation on 300 samples using BERT Precision and BERT Recall~\cite{zhang2019bertscore}, which measure semantic similarity between clean and adversarial outputs using contextual transformer embeddings. BERT Precision measures how well adversarial outputs preserve the semantic content of clean outputs, while BERT Recall measures how much semantic information from the clean outputs is retained after perturbation. We report both metrics under increasing perturbation budgets, where lower values indicate stronger adversarial degradation.
\color{black}

\noindent \textbf{Optimization details.}
Universal attacks optimize a single perturbation per model and attack configuration under an \(L_\infty\)-bounded perturbation radius \(c\), whereas sample-specific attacks optimize individual perturbations for each input. All attack objectives are optimized using Adam, which has been shown to achieve fast convergence with performance comparable to L-BFGS, under an \(L_\infty\)-projection constraint ~\cite{carlini2017towards}. Based on hyperparameter tuning, we use learning rates of \(10^{-4}\) for DiffAE, NVAE, TC-VAE, and \(\beta\)-VAE, \(10^{-2}\) for MAE, and \(10^{-3}\) for Gemma~3 and Qwen~2.5.

\subsection{Model conditioning behavior}
\label{sec:expConditioningBehaviour}
\begin{figure}[t]
    \centering
    \begin{minipage}[b]{0.23\textwidth}
        \centering
        \includegraphics[width=\linewidth]{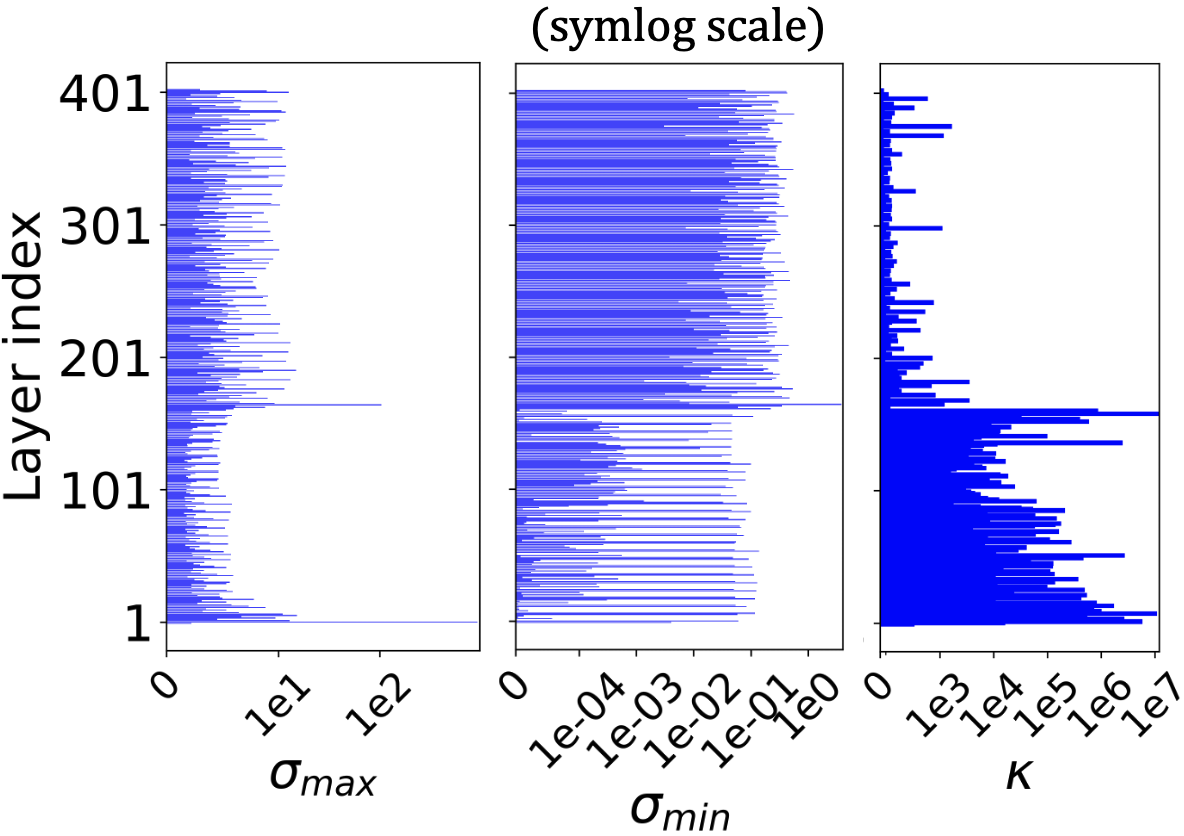}
        \caption*{Gemma 3}
    \end{minipage}\hfill
    \begin{minipage}[b]{0.23\textwidth}
        \centering
        \includegraphics[width=\linewidth]{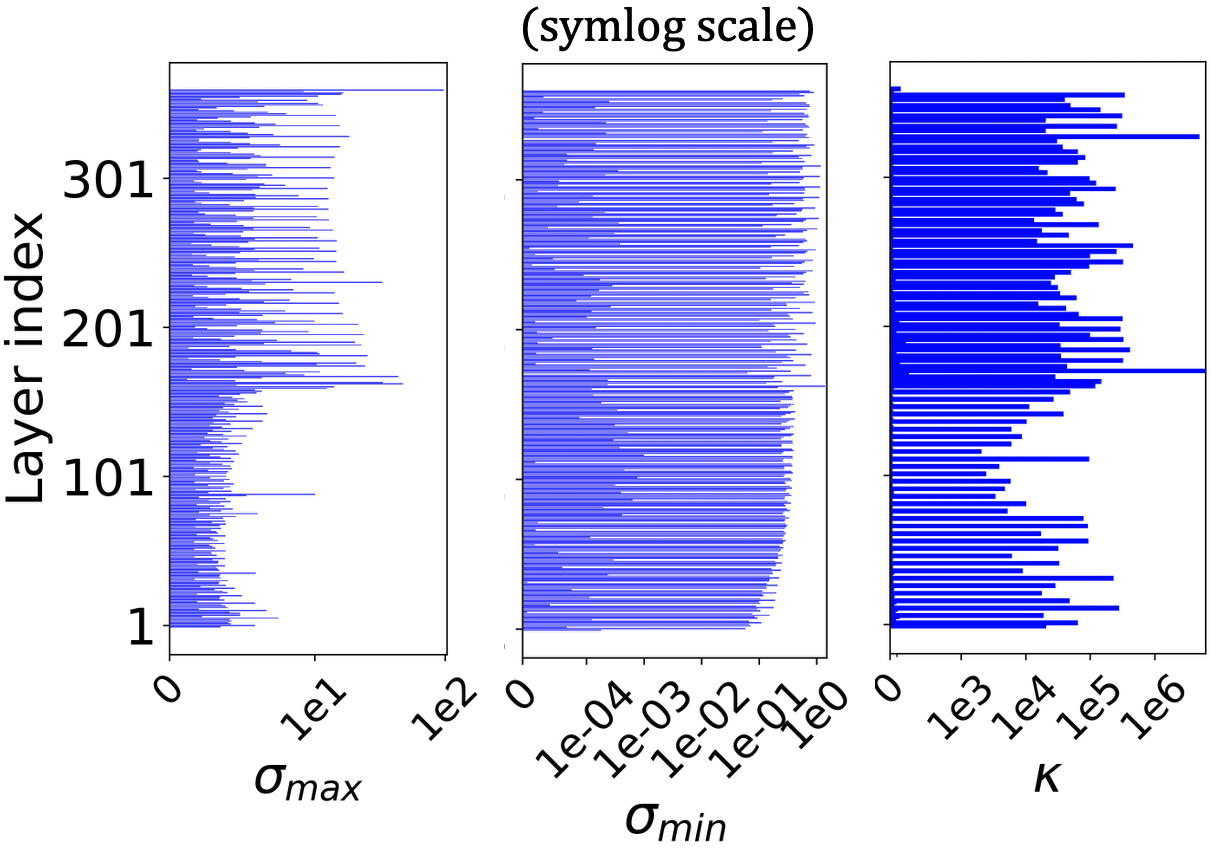}
        \caption*{Qwen 2.5}
    \end{minipage}
    \caption{Layer-wise conditioning profiles ($\sigma_{\max}$, $\sigma_{\min}$ and $\kappa$)}
    \label{fig:gemmaQwen}
\end{figure}

To understand the effectiveness of \ours{}, we first analyze the conditioning behavior of the evaluated models through the singular value spectra of intermediate transformations, using the maximum singular value $\sigma_{\max}$, minimum singular value $\sigma_{\min}$, and condition number $\kappa$. Following~\cite{yoshida2017spectral}, we compute approximate singular values and condition numbers of individual intermediate layer weights for each model. We exclude intermediate operations for which the condition number \(\kappa\) is not well defined. Large $\kappa$ values and near-zero $\sigma_{\min}$ indicate stronger ill-conditioning and an increased risk of gradient degradation.

\(\beta\)-VAE, TC-VAE, and MAE exhibit relatively mild ill-conditioning, as indicated by the absence of extremely small \(\sigma_{\min}\) values in intermediate layers (Figures~\ref{fig:tc_vae_s_s_k},~\ref{fig:beta_vae_s_s_k}, and~\ref{fig:mae_s_s_k}, respectively). NVAE (Figure~\ref{fig:nvae_s_s_k}) and the VLMs Gemma~3 and Qwen~2.5 (Figure~\ref{fig:gemmaQwen}) exhibit widespread ill-conditioning characterized by near-zero singular values and large condition numbers. DiffAE exhibits localized severe ill-conditioning concentrated in the final encoder layer (Figure~\ref{fig:diffae_s_s_k}).

Overall, the evaluated models span mild (\(\beta\)-VAE, TC-VAE, and MAE), localized (DiffAE), and severe ill-conditioning (NVAE, Gemma-3, Qwen~2.5) regimes, enabling systematic evaluation of \ours{} under varying levels of gradient degradation.

\subsection{Universal Attacks on AEs}
\label{sec:expUniversalAttacksAAEs}
We first evaluate GRILL in a \emph{classical universal attack setup}, where AE models operate without defenses. We then consider a stronger \emph{adaptive attack setup} in which a state-of-the-art defense mechanism is integrated into each AE model. 
In the classical setup, we compare \ours{} against the baselines OA and LA  under all attack configurations. In the adaptive setup, we evaluate the strongest baseline and \ours{} configurations identified in the classical setup.
\begin{figure}[t]
    \centering
    \begin{minipage}[b]{0.32\linewidth}
        \includegraphics[width=\linewidth]{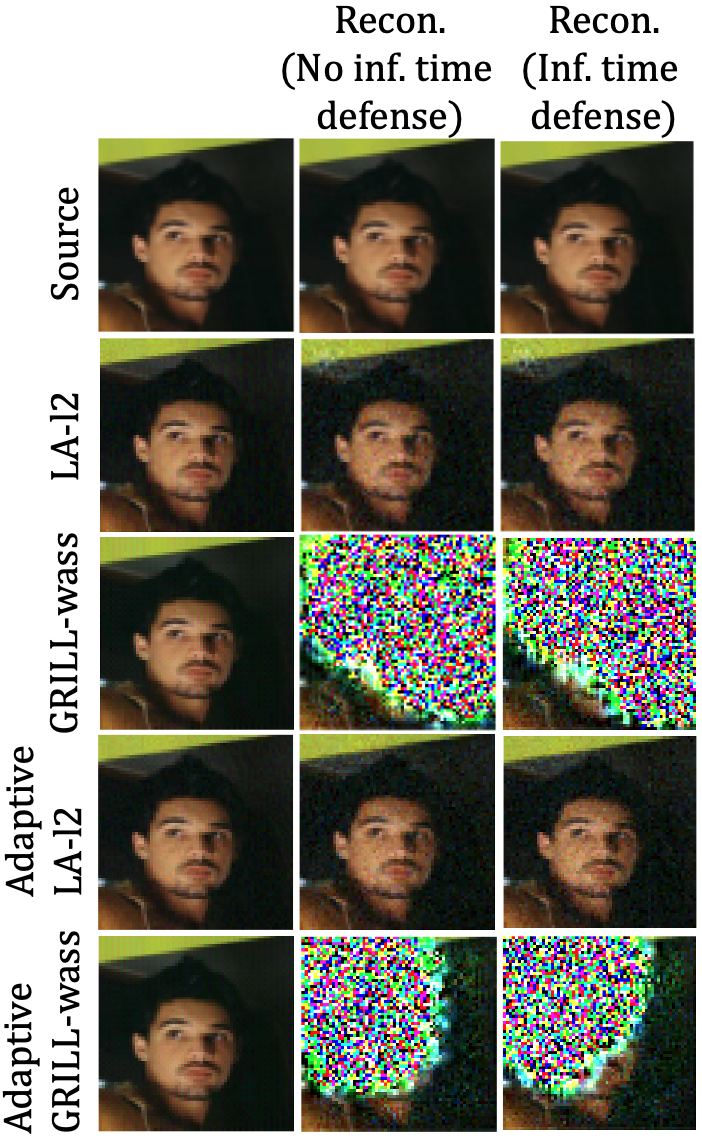}
        \caption*{NVAE (\(c=0.025\))}
    \end{minipage}
    \begin{minipage}[b]{0.32\linewidth}
        \includegraphics[width=\linewidth]{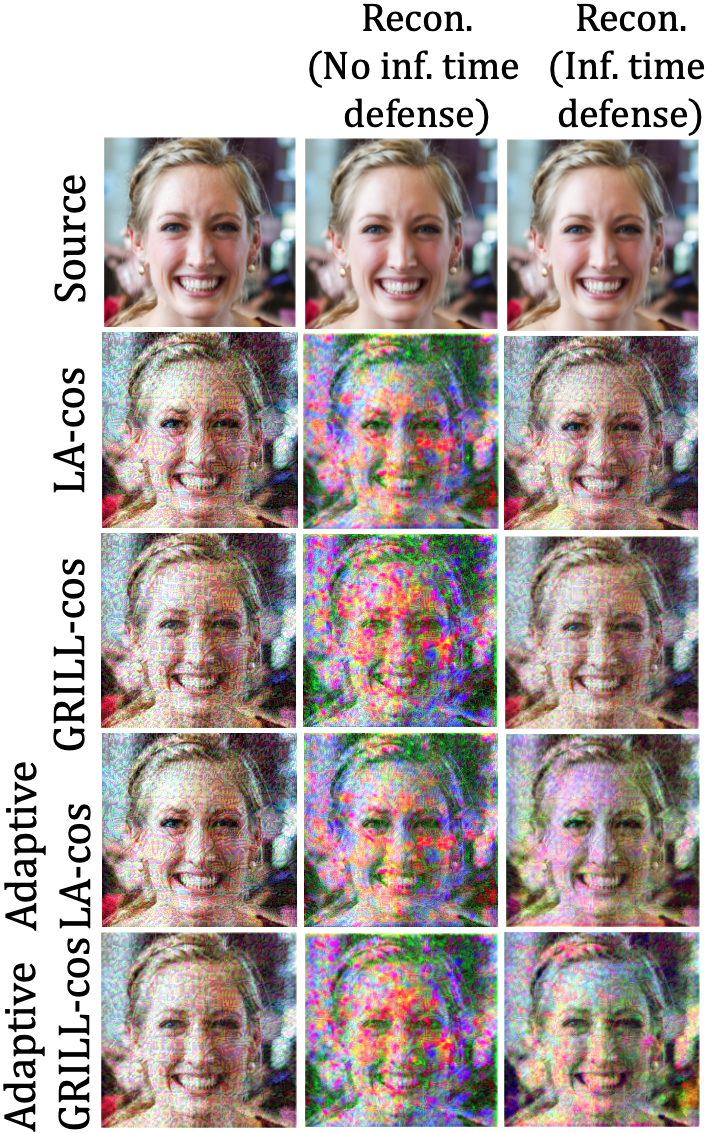}
        \caption*{DiffAE (\(c=0.25\))}
    \end{minipage}
    \begin{minipage}[b]{0.32\linewidth}
        \includegraphics[width=\linewidth]{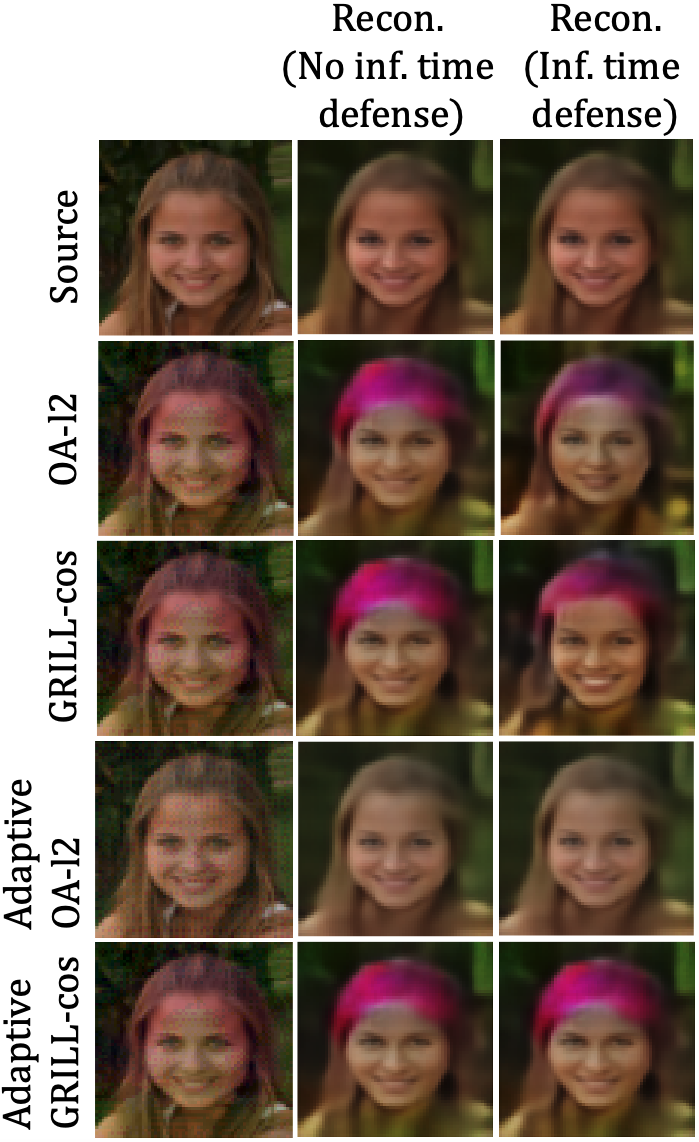}
        \caption*{TC-VAE (\(c=0.04\))}
    \end{minipage}
    \begin{minipage}[b]{0.32\linewidth}
        \includegraphics[width=\linewidth]{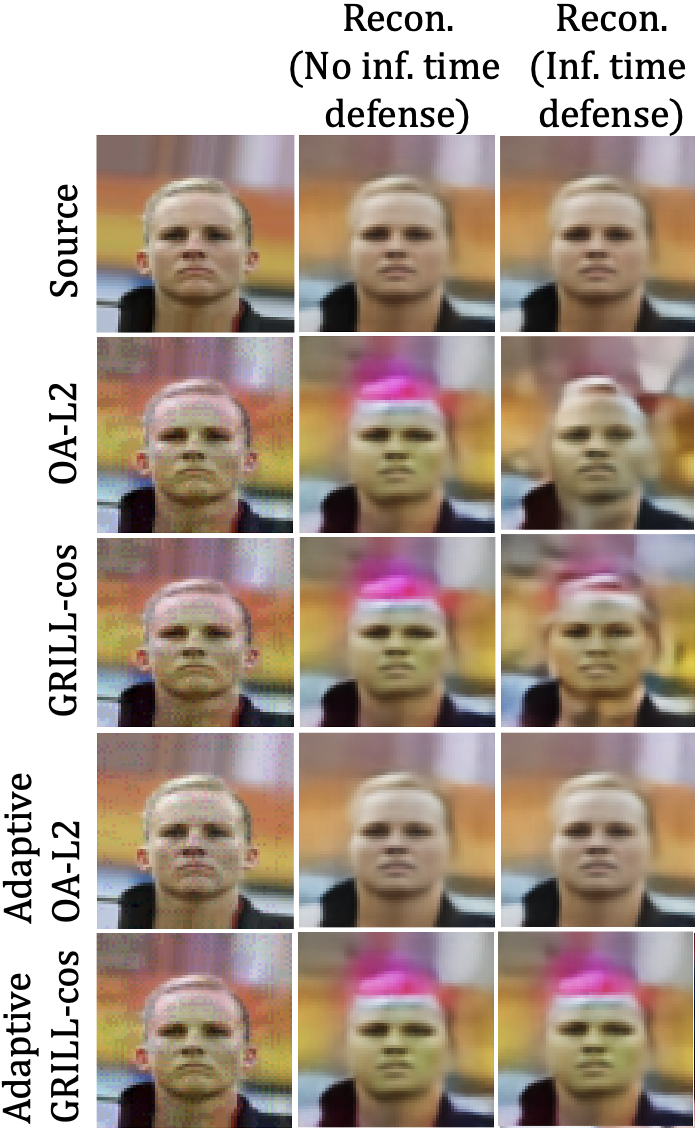}
        \caption*{\(\beta\)-VAE (\(c=0.04\))}
    \end{minipage}
    \begin{minipage}[b]{0.32\linewidth}
        \includegraphics[width=\linewidth]{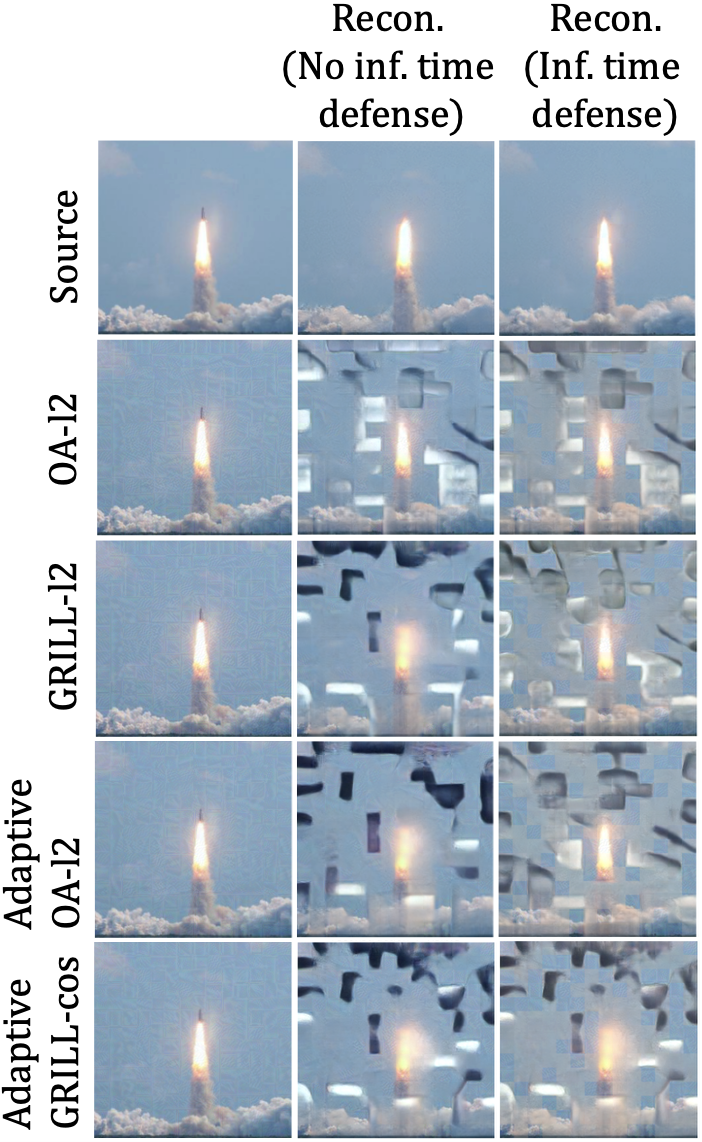}
        \caption*{MAE (\(c=0.09\))}
    \end{minipage}
    \caption{Universal attacks under HMC-based defense. Reconstructions are shown for baseline and \ours{} attacks with and without defense.}
    \label{fig:la_alma_uni_qualitative}
\end{figure}
\paragraph{Classical Universal Attacks.} 
For each AE model, we evaluate output distortion (OD) under increasing perturbation radii $c$ 
and relate attack effectiveness to the conditioning behavior (Section~\ref{sec:expConditioningBehaviour}).

$\beta$-VAE, TC-VAE, and MAE are \emph{mildly ill-conditioned} (Figures~\ref{fig:tc_vae_s_s_k}, \ref{fig:beta_vae_s_s_k} and \ref{fig:mae_s_s_k}). 
Consistent with this, \ours-based attacks perform comparably to the strongest baselines as seen in Figures~\ref{fig:beta_vae_var}, ~\ref{fig:mae_var} for $\beta$-VAE and MAE, respectively. Interestingly, despite the absence of very high $\kappa$ in TC-VAE, \ours{}-$L_2$ and \ours{}-cos outperform the strongest baseline OA-$L_2$ attack by up to \textbf{12.66\%}. This suggests that \ours{} provides benefits beyond alleviating gradient degradation.

For NVAE which exhibits \emph{severe ill-conditioning} 
(Figure~\ref{fig:nvae_s_s_k}), all \ours{}-based attacks substantially outperform the baselines (Figure~\ref{fig:nvae_var}), increasing output distortion from 
\textbf{38.11\%} to \textbf{56.66\%}.

 DiffAE exhibits \emph{localized severe ill-conditioning} in its final encoder layer (Figure~\ref{fig:diffae_s_s_k}).
Correspondingly, \ours{}-cos consistently outperforms all baselines across the entire perturbation range (Figure~\ref{fig:diffae_var}).
Cosine similarity emerges as the most effective attack metric for DiffAE, with \ours{}-cos exceeding the strongest baseline attack by \textbf{13.89\%} to \textbf{16.31\%} in output distortion.

\paragraph{Universal Adaptive Attacks}
\label{sec:expAdaptiveAttacks}
We next evaluate \ours{} in a universal adaptive attack setting, where the adversary explicitly accounts for the deployed defense mechanism during the attack generation. A defended AE is modeled as
$
\mathcal{D}(x) = g(\mathcal{Y}(x)),
$
where \(\mathcal{Y}\) is the AE and \(g\) is a latent-space defense mechanism.
In our setup, \(g\) corresponds to a Hamiltonian Monte Carlo (HMC)-based latent refinement defense~\cite{kuzina2022alleviating}, which iteratively adjusts latent representations toward high-likelihood regions at inference time to recover stable reconstructions from adversarial perturbations while largely preserving clean inputs. Accordingly, the attack objective becomes
\begin{equation}\label{eq.adapt_goal}
  x_a^{\star}=\arg\max\limits_{{x_a\in B^{p}_{c}(x)}}
  \Delta\bigl(\mathcal{D}(x_a),\mathcal{D}(x)\bigr)
\end{equation}

We select the strongest baseline and best \ours{} configuration from the classical setup and optimize attacks under active defense across varying \(L_\infty\) constraints. As shown in Figure~\ref{fig:adaptive}, \ours{} consistently achieves higher distortion scores than the strongest baselines, with gains up to \textbf{101.99\%} for NVAE, \textbf{15.30\%} for DiffAE, \textbf{34.96\%} for TC-VAE, and \textbf{77.59\%} for \(\beta\)-VAE (Figure~\ref{fig:adaptive}). 
The gains of \ours{} become more pronounced for TC-VAE and \(\beta\)-VAE under adaptive attacks, despite comparatively smaller gains in the classical setting, because the HMC defense suppresses baseline attacks more strongly by operating directly in latent space. Although adaptive attacks produce lower overall distortion, \ours{} remains effective because, even when latent gradients vanish, the loss continues to influence adversarial updates by scaling non-vanishing gradients from earlier intermediate layers carrying adversarial signal. In contrast, no gains are observed for MAE because the HMC defense is less effective for its deterministic latent space.
\color{black}
\paragraph{Qualitative Analysis}
\label{sec:expAdaptiveAttacks}
Figure~\ref{fig:la_alma_uni_qualitative} shows that \ours{} induces stronger output distortions than the strongest baseline attacks for NVAE and DiffAE, particularly under HMC-based defense.
This effect is most pronounced for NVAE, consistent with its severe and widespread ill-conditioning. 
In contrast, TC-VAE, $\beta$-VAE, and MAE\ exhibit qualitatively similar behavior between baseline and \ours{}-based attacks, consistent with their milder conditioning profiles.

Inference-time defenses substantially suppress baseline attack distortions across all models, whereas GRILL-based attacks remain comparatively effective (Figure~\ref{fig:la_alma_uni_qualitative}: NVAE, DiffAE, TC-VAE, $\beta$-VAE). 
Although \ours{} does not outperform baseline attacks on MAE on average (Figure~\ref{fig:mae_var}), Figure~\ref{fig:la_alma_uni_qualitative}(MAE) shows that it can still induce stronger distortions for certain samples, suggesting that GRILL remains complementary even in comparatively well-conditioned settings.\color{black}

\begin{figure}[t]
    \centering
    \begin{minipage}[b]{0.8\linewidth}
        \centering
        \includegraphics[width=\linewidth]{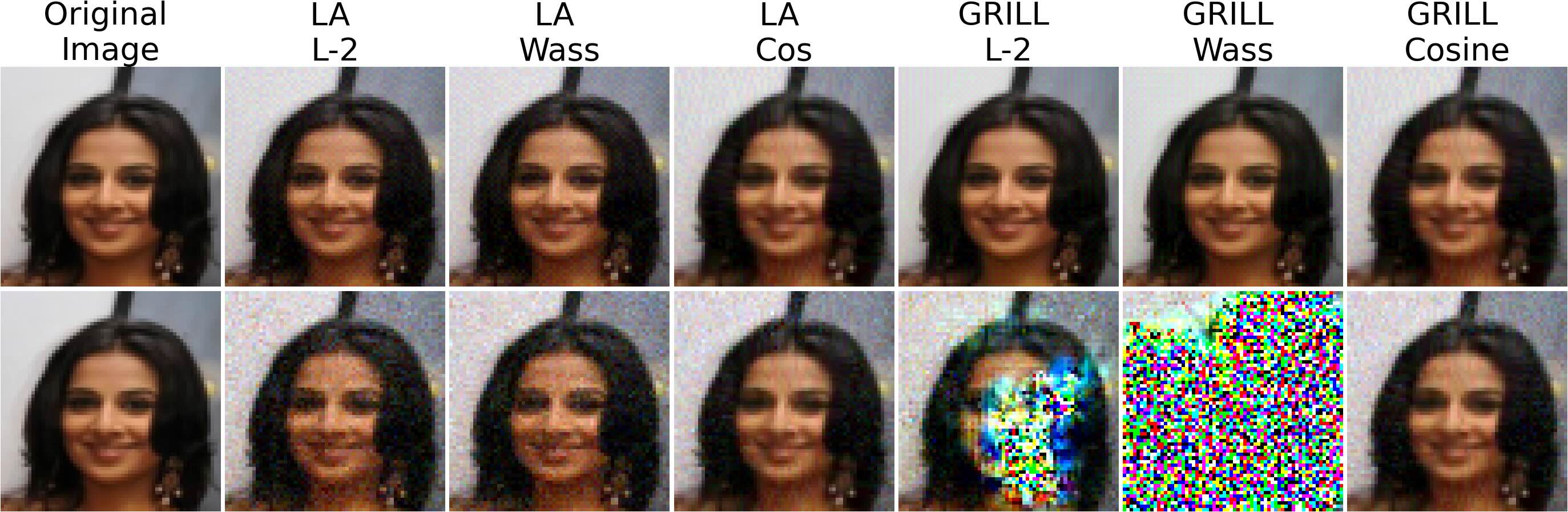}
        \includegraphics[width=\linewidth]{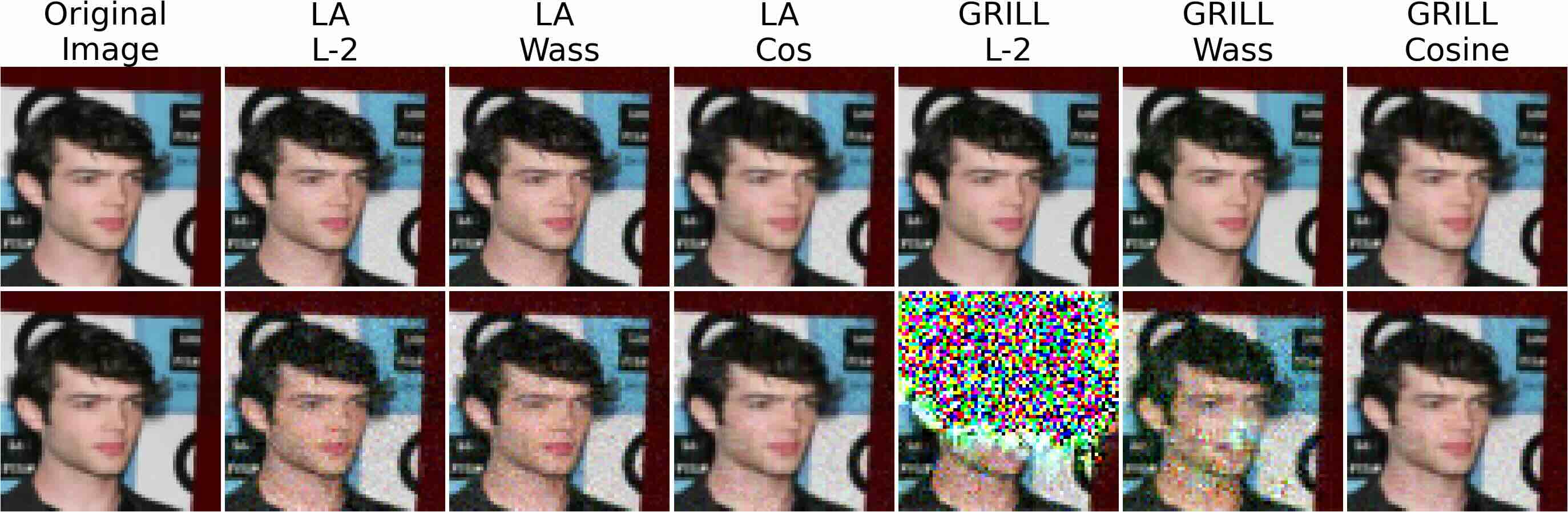}
        \includegraphics[width=\linewidth]{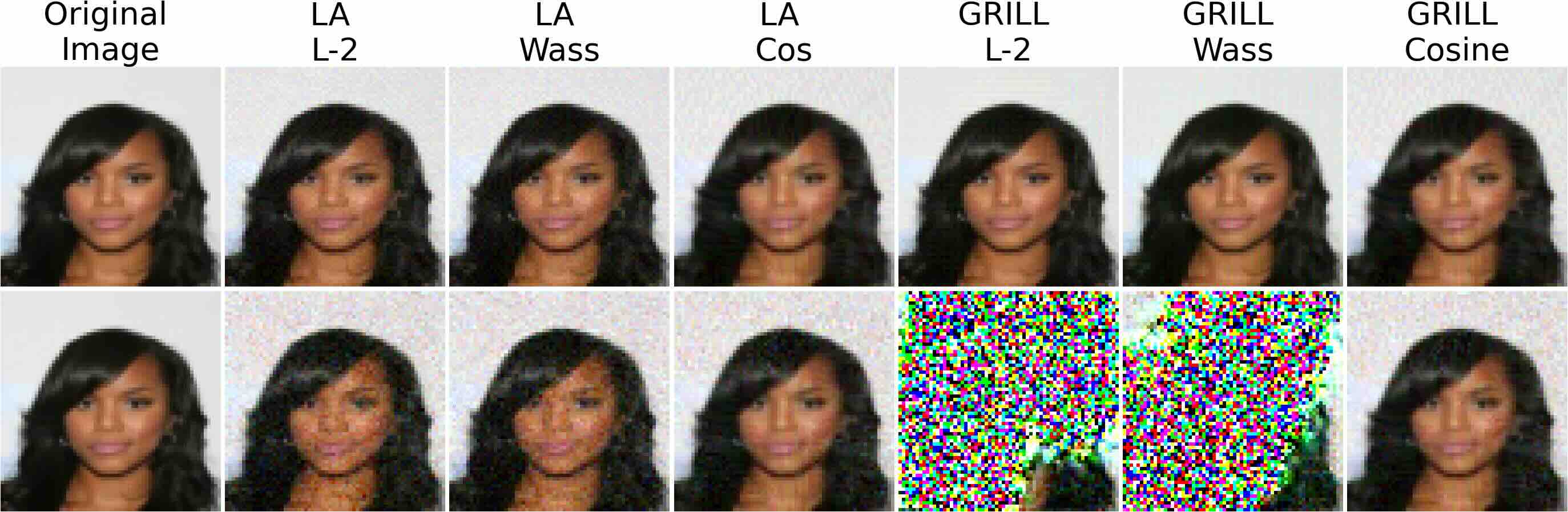}
        \label{fig:nvaeSampleSpecificAppendix}
        \caption*{(a) NVAE (\(c=0.03\))}
    \end{minipage}
    \vspace{0.5em}

    \begin{minipage}[b]{0.8\linewidth}
        \centering
        \includegraphics[width=\linewidth]{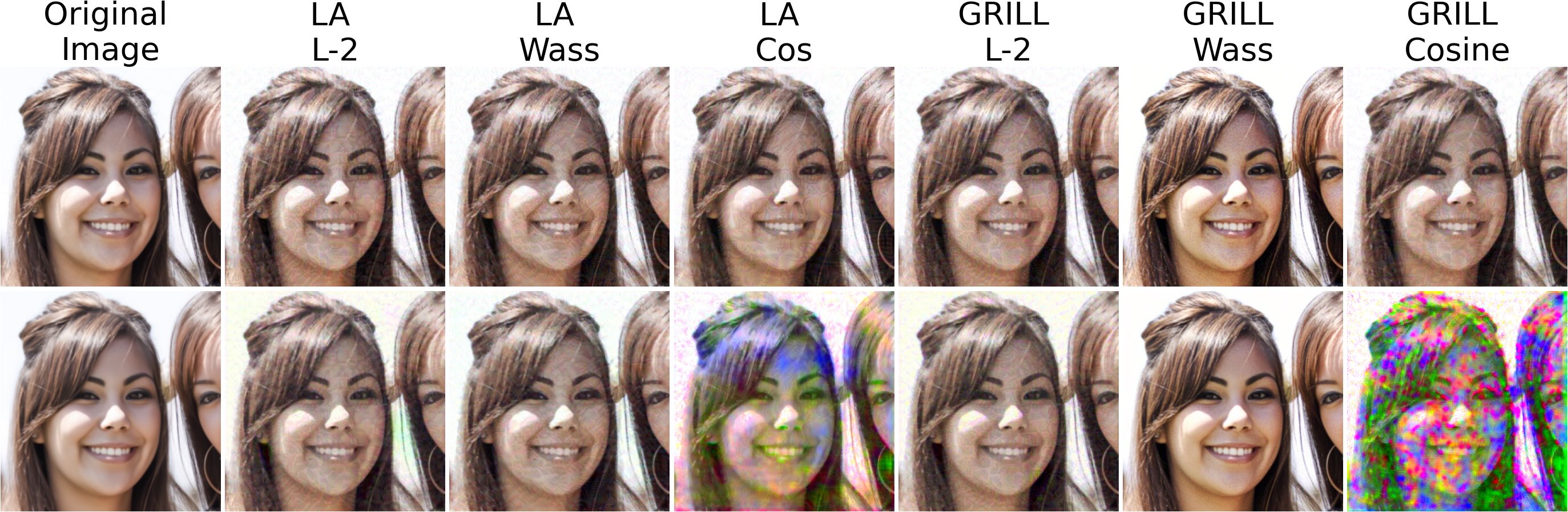}
        \includegraphics[width=\linewidth]{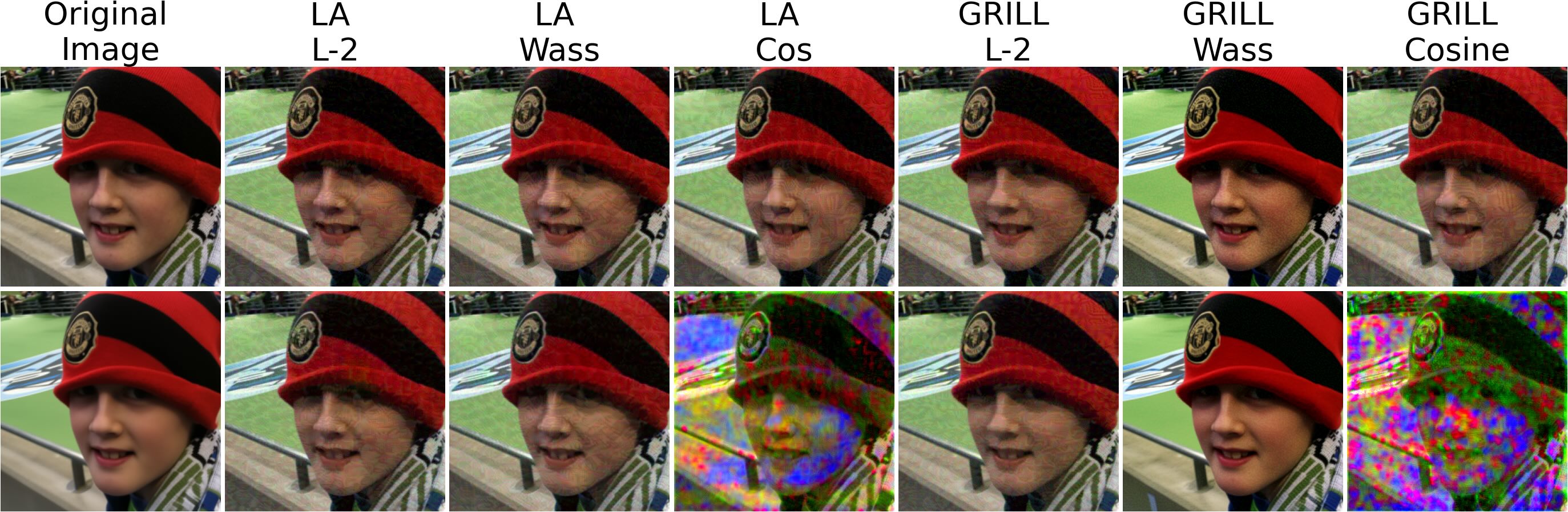}
        \includegraphics[width=\linewidth]{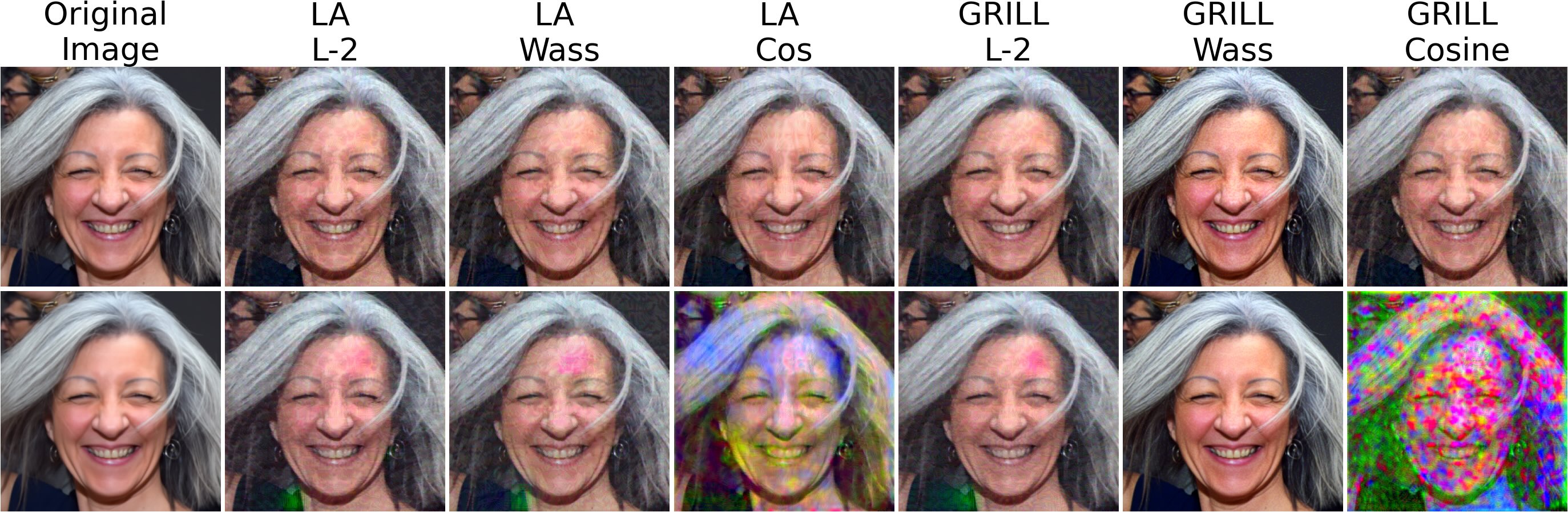}
        \label{fig:diffAESampleSpecificAppendix}
        \caption*{(b) DiffAE (\(c=0.08\))}
    \end{minipage}
    \caption{
    Sample-specific attacks on highly ill-conditioned models. Clean/adversarial inputs are shown on the top, with corresponding reconstructions on the bottom.
    }
    \label{fig:NVAE_DIffAE_sample_specific}
\end{figure}

\subsection{Sample-specific Attacks on AEs}
\label{sec:expSampleSpecificAttackesAE} 
Figures~\ref{fig:NVAE_DIffAE_sample_specific}(a) and~\ref{fig:NVAE_DIffAE_sample_specific}(b) present qualitative sample-specific attacks on the highly ill-conditioned models NVAE and DiffAE. Under imperceptible perturbations, \ours{} induces substantially stronger reconstruction distortions than the baseline LA attacks across multiple distance functions. Baseline attacks often fail to produce noticeable output degradation. These qualitative results support the conditioning analysis in Section~\ref{sec:expConditioningBehaviour} and further demonstrate the effectiveness of \ours{} in ill-conditioned AEs.
\color{black}
\subsection{Beyond AEs: Adversarial Attacks on Modern  Encoder–Decoder Architectures}
\label{sec:expAttacksVLMs}

\begin{figure*}[t]
    \centering
    \begin{minipage}[b]{1.0\textwidth}
        \centering
       \includegraphics[width=1.0\linewidth]{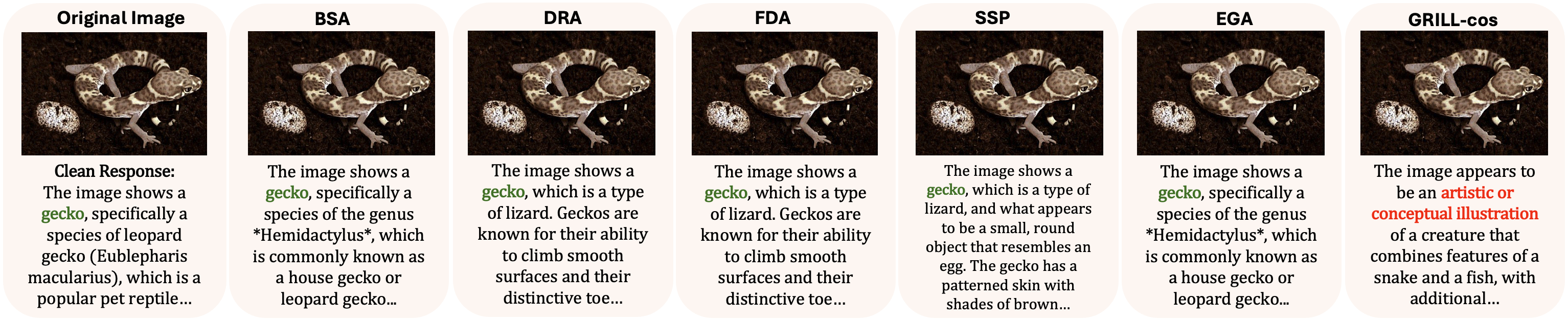}
        \caption*{ Qwen 2.5  ($c=0.002$)}
    \end{minipage}\hfill
    \begin{minipage}[b]{1.0\textwidth}
        \centering
       \includegraphics[width=1.0\linewidth]{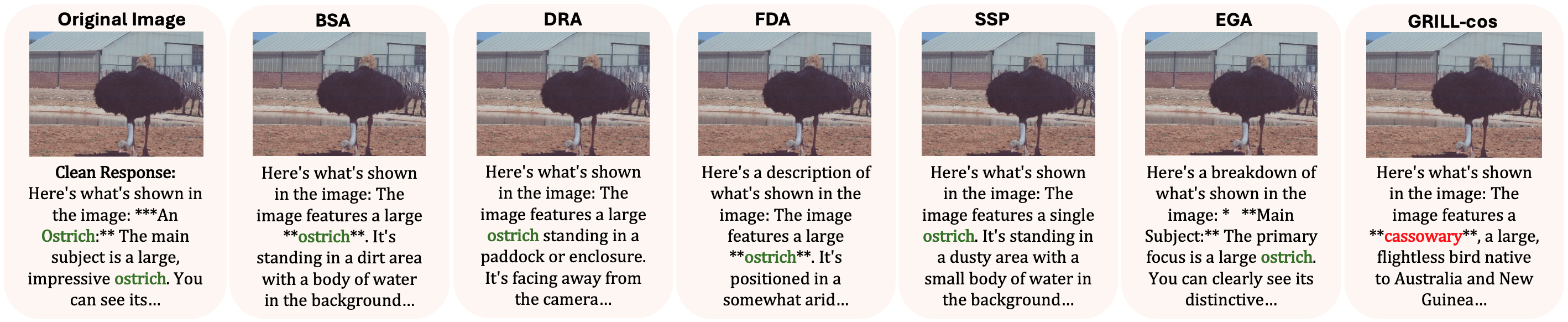}
        \caption*{ Gemma 3 ($c=0.0006$)}
    \end{minipage}\hfill
    \caption{Adversarially perturbed images evaluated using the prompt \textbf{``What is shown in this image?''} under small perturbation budgets $c$. Ellipses (…) indicate omitted text without loss of context.}
    \label{fig:QwenGemmaQualitative}
\end{figure*}

We evaluate \ours{} on modern multimodal encoder–decoder architectures, Gemma 3 and Qwen 2.5. In this setting, the output logits of the VLM are treated as the attack output space, while intermediate hidden states serve as latent feature representations. Modern VLMs similarly employ layer-wise latent representations and encoder–decoder-style multimodal transformations \cite{tan2019lxmert,li2022blip}, motivating investigation of \ours{} beyond classical AEs. Among the evaluated \ours{} distance configurations, \ours{}-cos consistently achieved the strongest attack performance; therefore, we report only GRILL-cos results for VLM experiments. We compare \ours{}-cos against several representative white-box attack baselines, including DRA, FDA, SSPA, BSA, and EGA (see Section~\ref{sec:related_work}). Quantitative evaluation is reported for Gemma 3, while for the computationally more expensive Qwen 2.5 model, we report qualitative results only, as obtaining statistically significant quantitative results would require expensive evaluation.

\noindent \textbf{Qualitative results:}
Figure \ref{fig:QwenGemmaQualitative} shows that under small perturbation budgets ($c \leq 0.002$ for Gemma 3 and $c \leq 0.0006$ for Qwen~2.5), baseline attacks induce only minor semantic variations in the generated response, indicating limited attack effectiveness. In contrast, \ours{}-cos produces substantially stronger semantic degradation, including contextual distortion and loss of object fidelity.


This behavior is consistent with the conditioning analysis (Figure~\ref{fig:gemmaQwen}), where both models exhibit widespread intermediate-layer ill-conditioning characterized by near-zero singular values. The results suggest that \ours{} remains effective on modern multimodal architectures even under highly constrained perturbation budgets.

\begin{figure}[t]
    \centering
    \begin{minipage}[b]{0.48\linewidth}
        \includegraphics[
            width=\linewidth,
            trim={0.1cm 0.2cm 0.3cm 0.2cm},
            clip
        ]{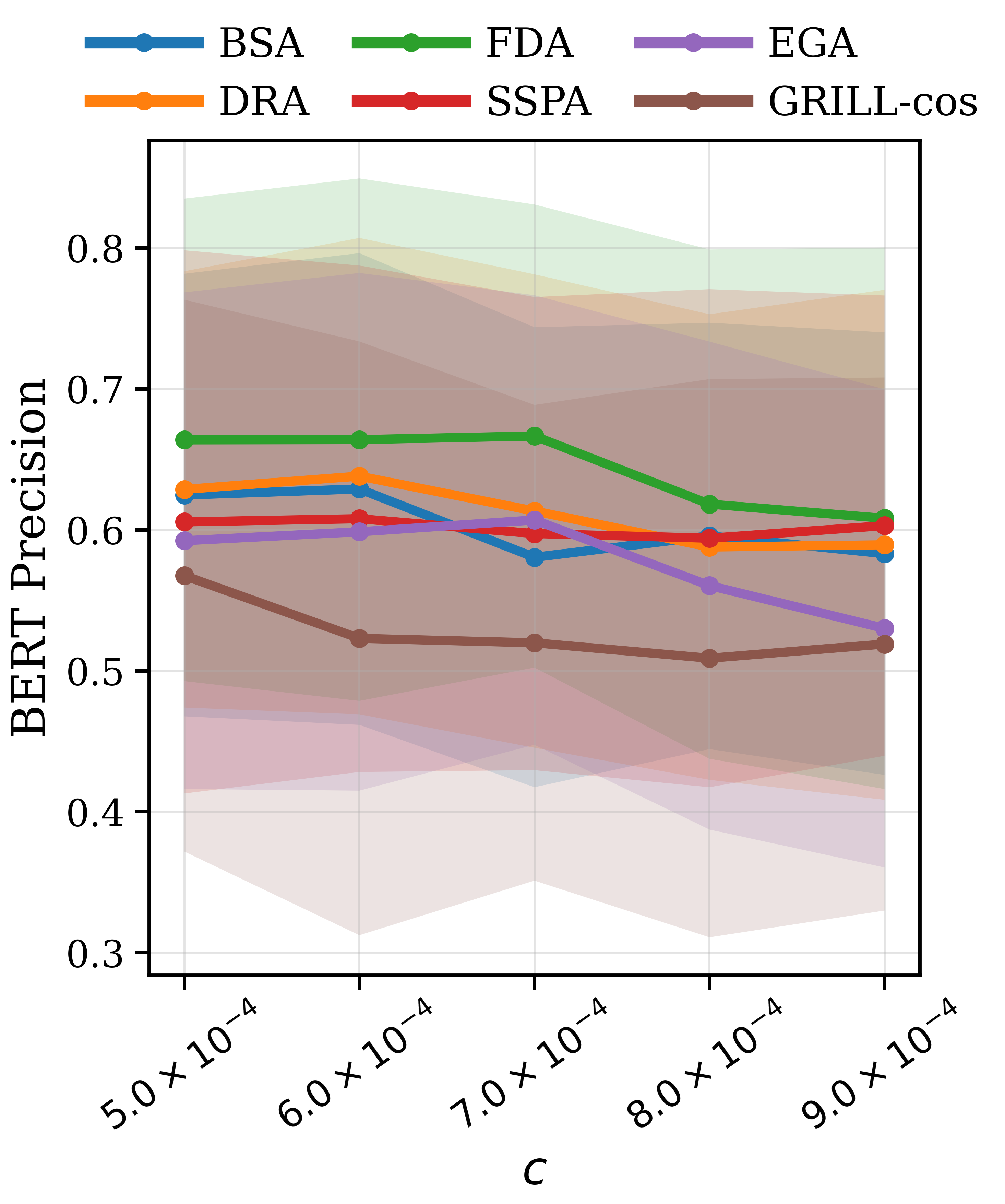}
        \caption*{(a)}
    \end{minipage}
    \begin{minipage}[b]{0.48\linewidth}
        \includegraphics[
            width=\linewidth,
            trim={0.1cm 0.2cm 0.3cm 0.2cm},
            clip
        ]{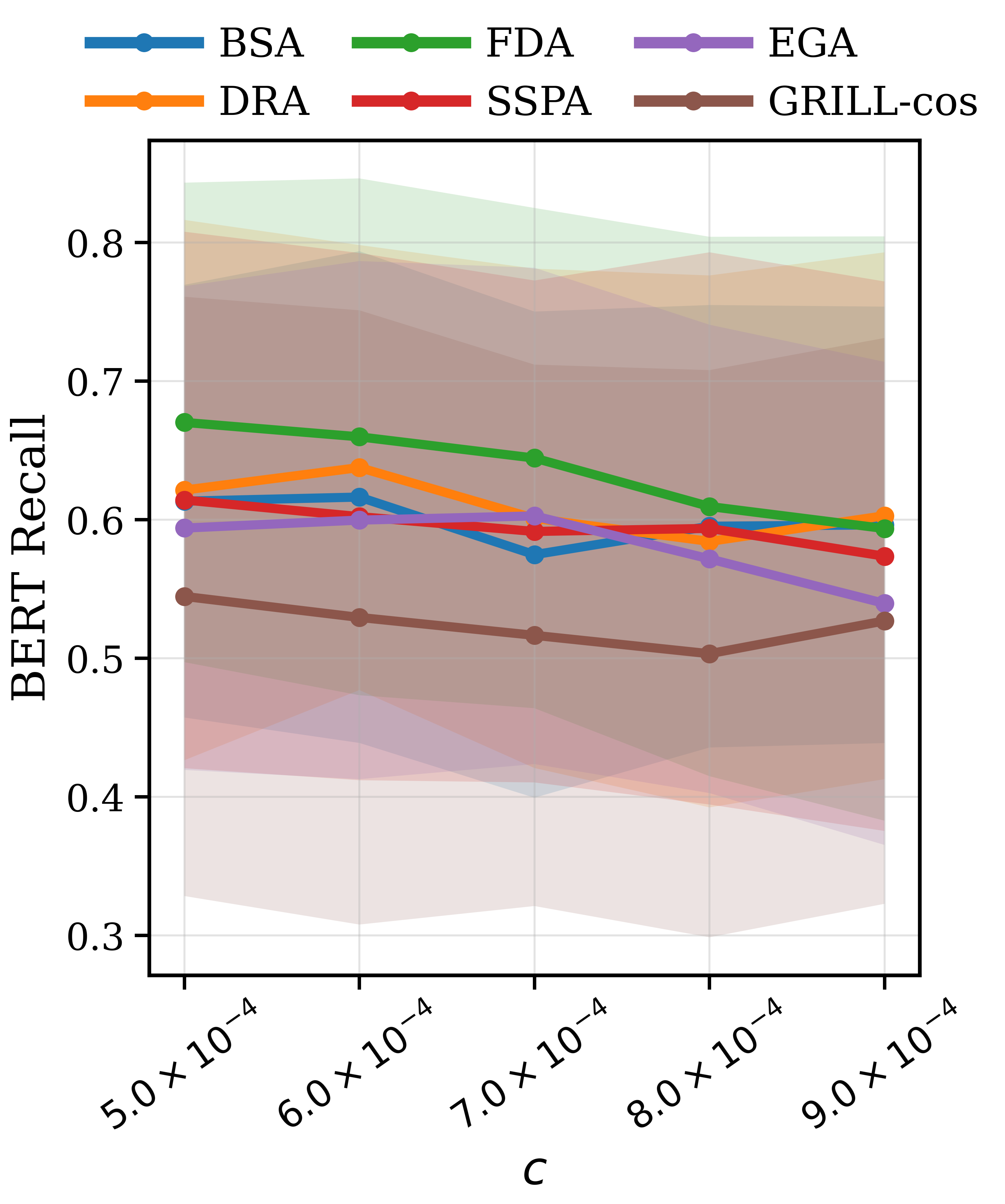}
        \caption*{(b)}
    \end{minipage}
    \caption{Gemma 3 quantitative results of sample--specific attacks}
    \label{fig:gemma3Quant}
\end{figure}

\noindent \textbf{Quantitative results:}
Figure \ref{fig:gemma3Quant} compares attack performance under increasing perturbation budgets using BERT Precision and Recall.
Across all evaluation metrics, \ours{}-cos consistently achieves lower scores than BSA, DRA, FDA, SSPA, and EGA, indicating stronger semantic degradation and more effective attacks.

\emph{Note:} All qualitative results are shown for smaller perturbation budgets, where gradient degradation is prevalent and \ours{} clearly outperforms baseline attacks. At higher budgets, output distortions from baseline attacks become comparable.

\subsection{Ablation study, Convergence \& Efficiency}
\label{sec:expAblationEfficiency}
We further analyze the behavior of \ours{} through ablation, convergence, and efficiency studies. 

\begin{figure}[t]
    \centering
    \begin{minipage}[b]{0.4\linewidth}
        \includegraphics[width=\linewidth]{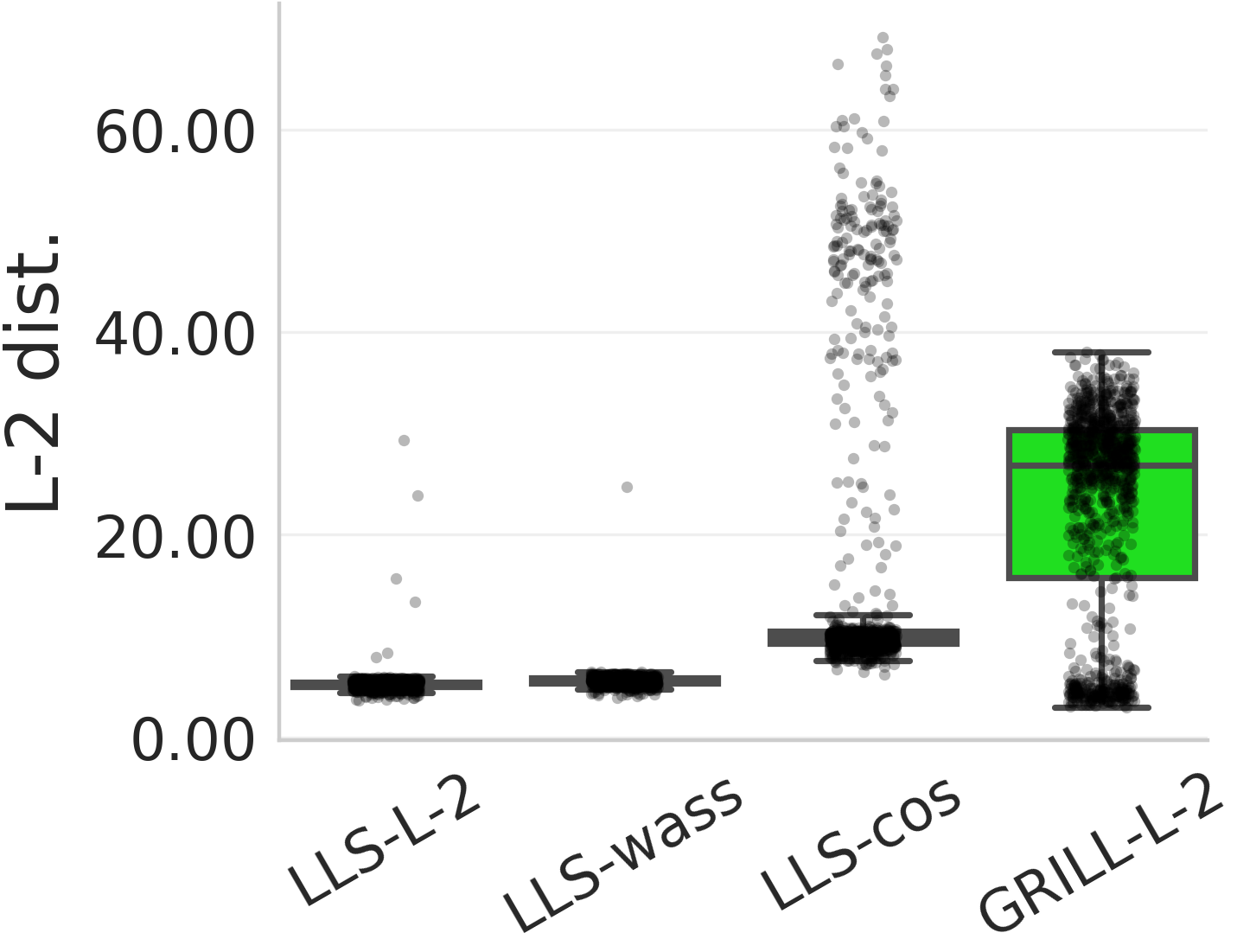}
        \caption*{NVAE ($c=0.025$)}
    \end{minipage}
    \begin{minipage}[b]{0.4\linewidth}
        \includegraphics[width=\linewidth]{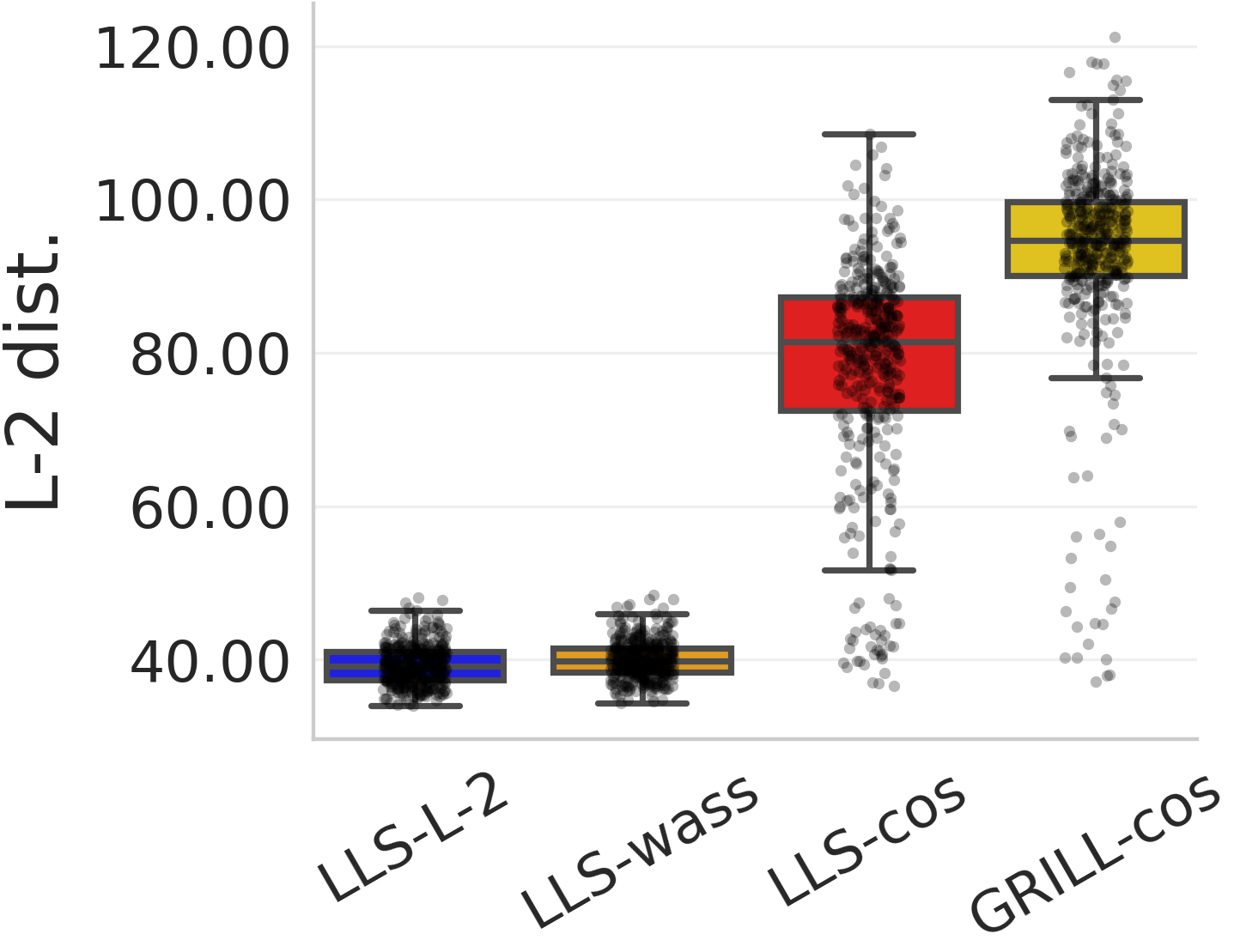}
        \caption*{DiffAE ($c=0.22$)}
    \end{minipage}
    \caption{Ablation: \ours{} vs layer-wise loss aggregation (LLS)}
    \label{fig:aggreAblation}
\end{figure}

\noindent\textbf{GRILL vs Layer-wise loss Summation (LLS):}
Figure~\ref{fig:aggreAblation} compares \ours{} against a variant obtained by removing the distortion term \(\delta^*\) from Eq.~\ref{eq.grill_final}, reducing the objective to simple layer-loss summation (LLS). 
We evaluate both formulations under the base perturbation budgets (\(c=0.025\) for NVAE and \(c=0.22\) for DiffAE, see Table~\ref{tab:models_datasets_radii}), where gradient degradation effects are most pronounced. 

\ours{} consistently outperforms LLS, with particularly large gains on NVAE (Figure~\ref{fig:aggreAblation}) due to its widespread ill-conditioning (Figure~\ref{fig:nvae_s_s_k}) across multiple intermediate layers. Despite exhibiting only localized severe ill-conditioning (Figure ~\ref{fig:diffae_s_s_k}), DiffAE  also shows a substantial performance gap (Figure~\ref{fig:aggreAblation}: DiffAE). These results indicate that simple layer-wise loss aggregation is insufficient to mitigate gradient degradation effectively and further support the hypothesis. 
Additionally, Figure~\ref{fig:gemma3Quant} shows that \ours{} outperforms BSA, which is based on layer-wise and token-wise cosine similarity aggregation, further indicating the superiority of \ours{} over conventional aggregation-based attack formulations that lack the $\delta^*$ term from Eq.~\ref{eq.grill_final}.

\begin{figure}[t]
    \centering
    \begin{subfigure}[b]{0.23\textwidth}
        \centering
        \includegraphics[
            width=\linewidth,
            height=4cm,
            trim=0.5cm 0.4cm 0.5cm 0.6cm,
            clip
        ]{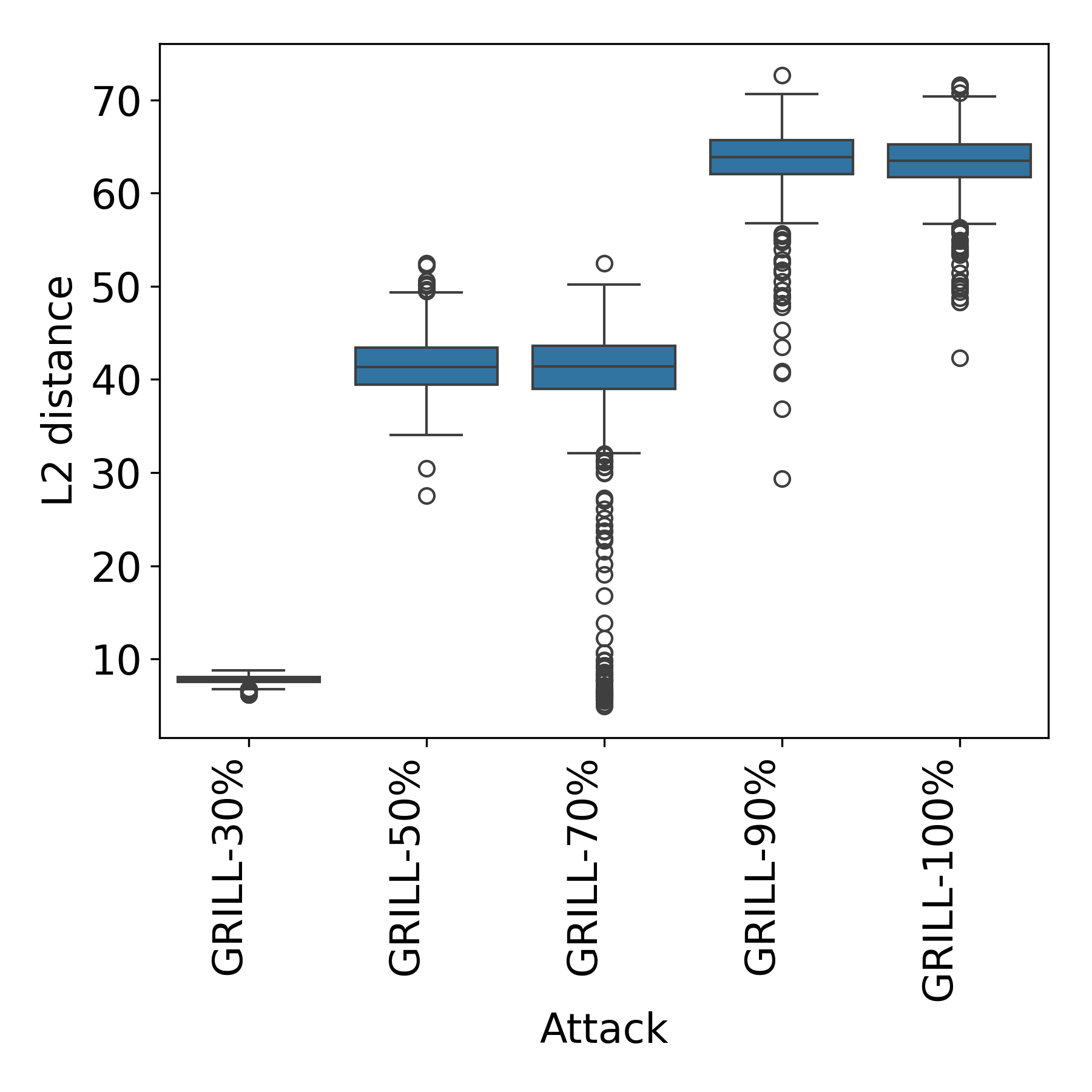}
    \end{subfigure}\hfill
    \caption{Output distortion distributions for split-wise \ours{} ablations on NVAE under varying fractions of encoder--decoder splits.}
    \label{fig:ablbox}
\end{figure}

\noindent\textbf{Effect of the Fraction of Encoder--Decoder Splits Used in GRILL:} To analyze the role of partial split aggregation, we vary the fraction of encoder--decoder splits considered in Equation~\ref{eq.grill_final} from 30\% to 100\% using \ours{}-wass on NVAE, which serves as an ideal testbed due to its severe ill-conditioning.
As shown in Figure~\ref{fig:ablbox}, increasing the proportion of  intermediate splits consistently improves attack strength, leading to larger output distortions.

\begin{figure}[t]
    \centering
    \begin{subfigure}[b]{0.45\textwidth}
        \centering
        \includegraphics[width=\linewidth]{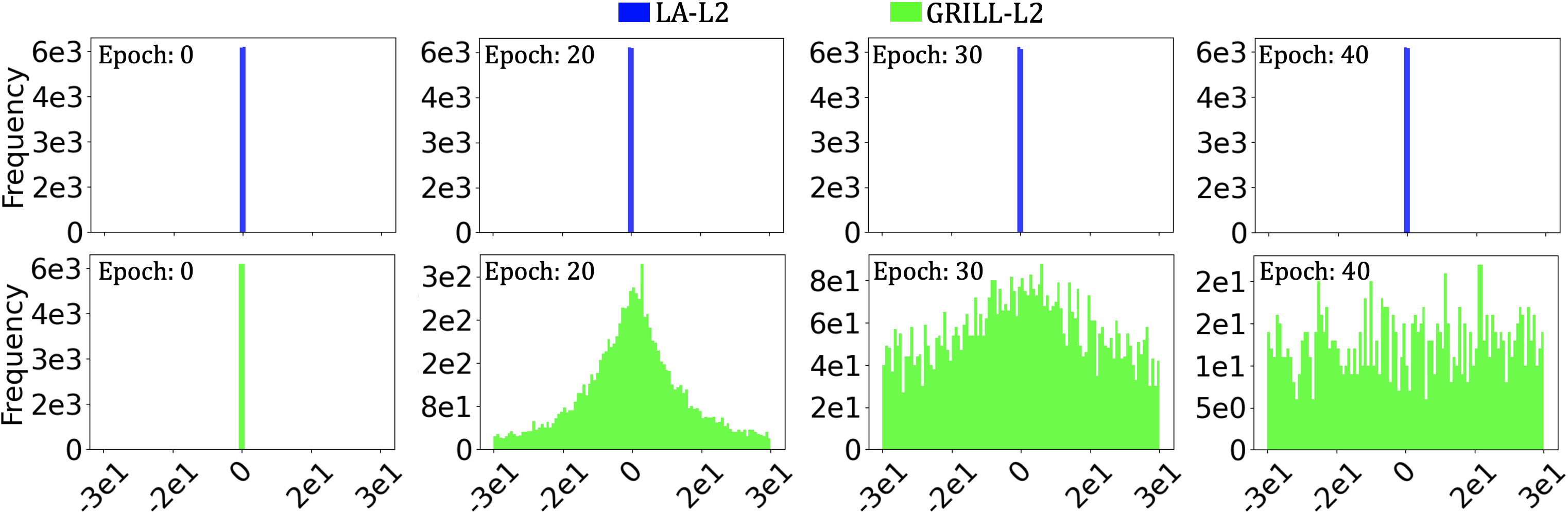}
        \caption{NVAE ($c = 0.08$)}
    \end{subfigure}
    \hfill
    \begin{subfigure}[b]{0.45\textwidth}
        \centering
        \includegraphics[width=\linewidth]{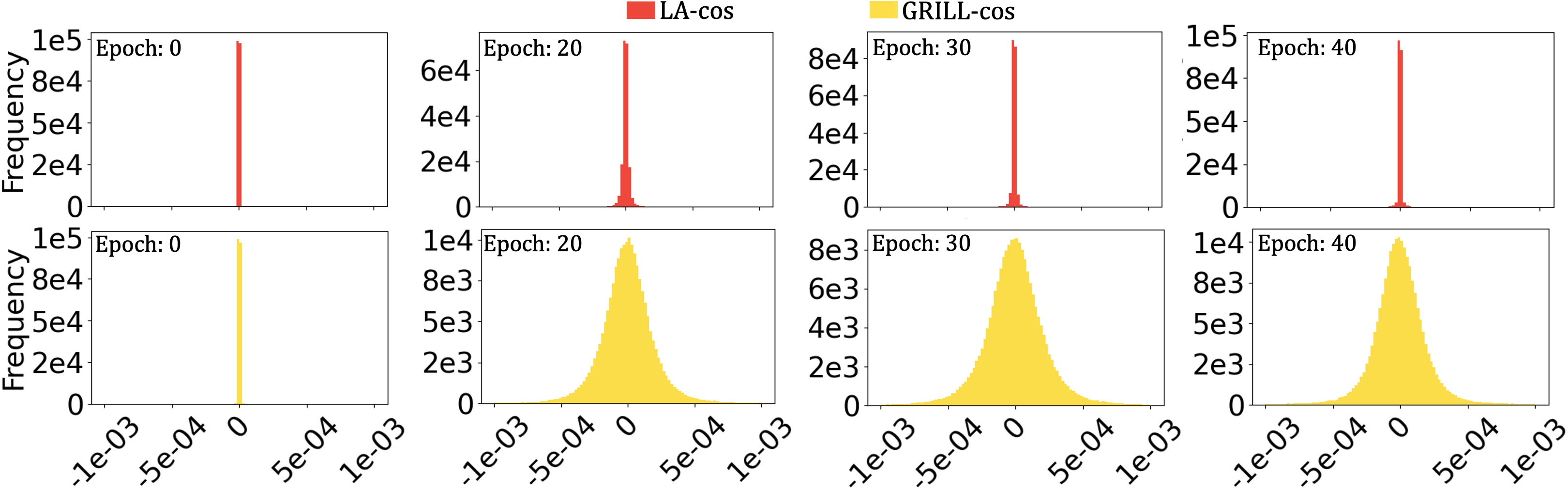}
        \caption{DiffAE ($c = 0.3$)}
    \end{subfigure}
    \caption{Adversarial loss gradient histograms for baseline and \ours{}-based universal attacks.}
    \label{fig:diffHist}
\end{figure}

\noindent\textbf{Gradient degradation and recovery:}
To validate our hypothesis that weak attacks arise from gradient degradation,
we compare histograms of backpropagated adversarial loss gradients for baseline attacks and \ours{} during optimization. As shown in Figure~\ref{fig:diffHist}, baseline gradients remain sharply concentrated near zero throughout optimization, indicating severe gradient degradation.
In contrast, \ours{} produces substantially broader, lower-kurtosis gradient distributions with larger magnitudes, indicating improved adversarial gradient propagation during optimization. The behavior is consistently observed for both NVAE and DiffAE. 

\begin{figure}[t]
    \centering
\includegraphics[width=1.0\linewidth]{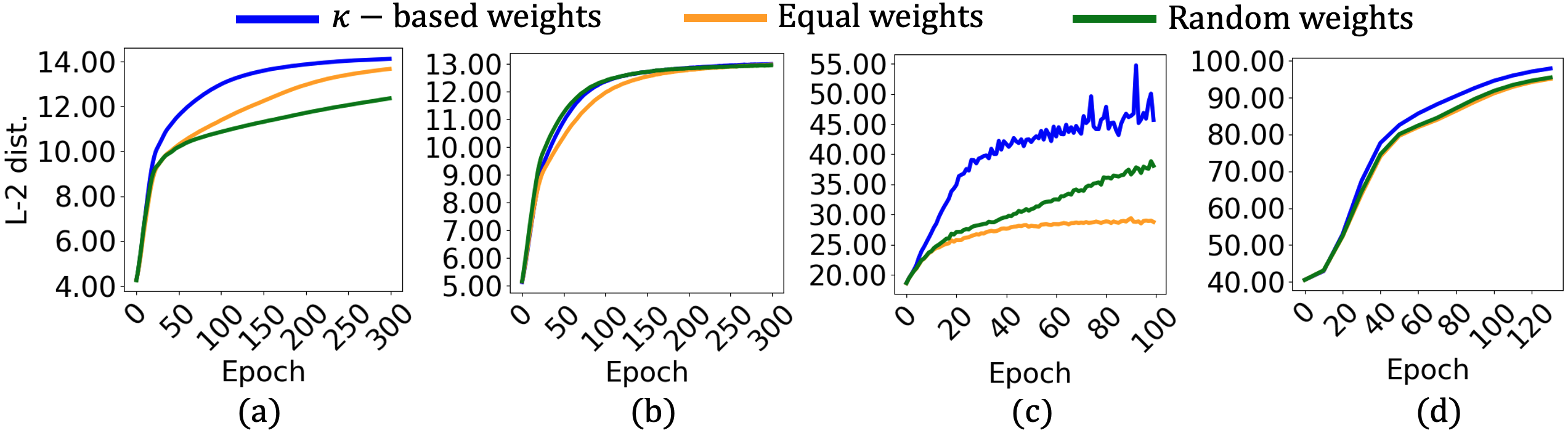}
\caption{Equal, random, and $\kappa$-based weighting strategies for \ours{} across (a) $\beta$-VAE, (b) TC-VAE, (c) NVAE (d) DiffAE.} \label{fig:ablation}
\end{figure}

\noindent\textbf{Weighting strategy:} 
We investigate conditioning-aware weighting strategies for \ours{}’s intermediate split losses. Specifically, we compare equal, random, and condition-number-based ($\kappa$-based) weighting schemes, in which more ill-conditioned split interfaces (larger $\kappa$) are assigned lower weights, whereas better-conditioned split interfaces receive higher weights. As shown in Figures~\ref{fig:ablation}(a–d), $\kappa$-based weights noticeably impact optimization, particularly in ill-conditioned models like NVAE (Figure \ref{fig:ablation}c).  In particular, $\kappa$-based weighting improves optimization behavior for highly ill-conditioned models, suggesting that conditioning-aware weighting can further facilitate adversarial gradient propagation.

\begin{figure}[t]
    \centering
    \begin{minipage}[b]{0.32\linewidth}
        \includegraphics[width=\linewidth]{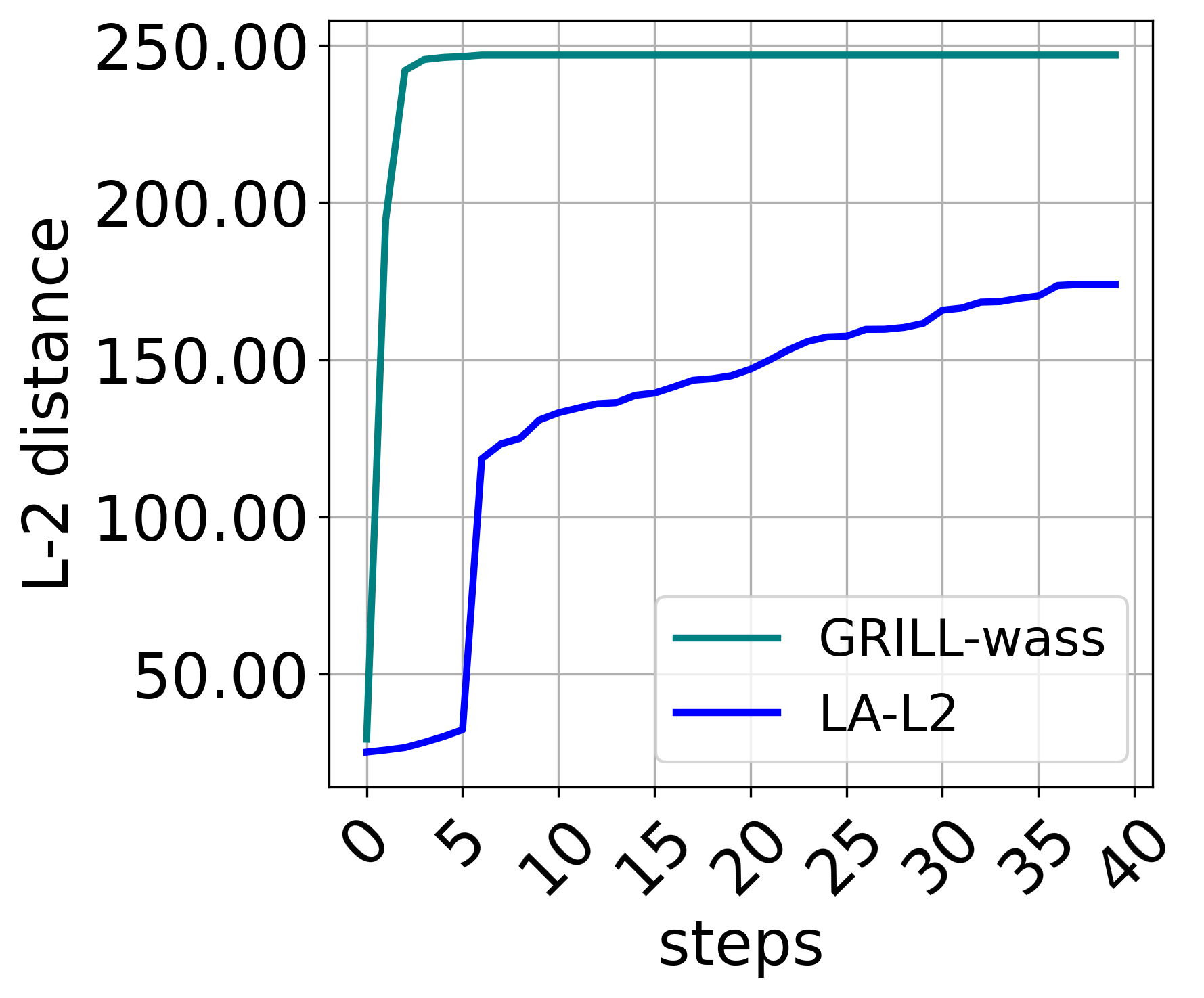}
        \caption*{(a) NVAE}
    \end{minipage}
    \begin{minipage}[b]{0.32\linewidth}
        \includegraphics[width=\linewidth]{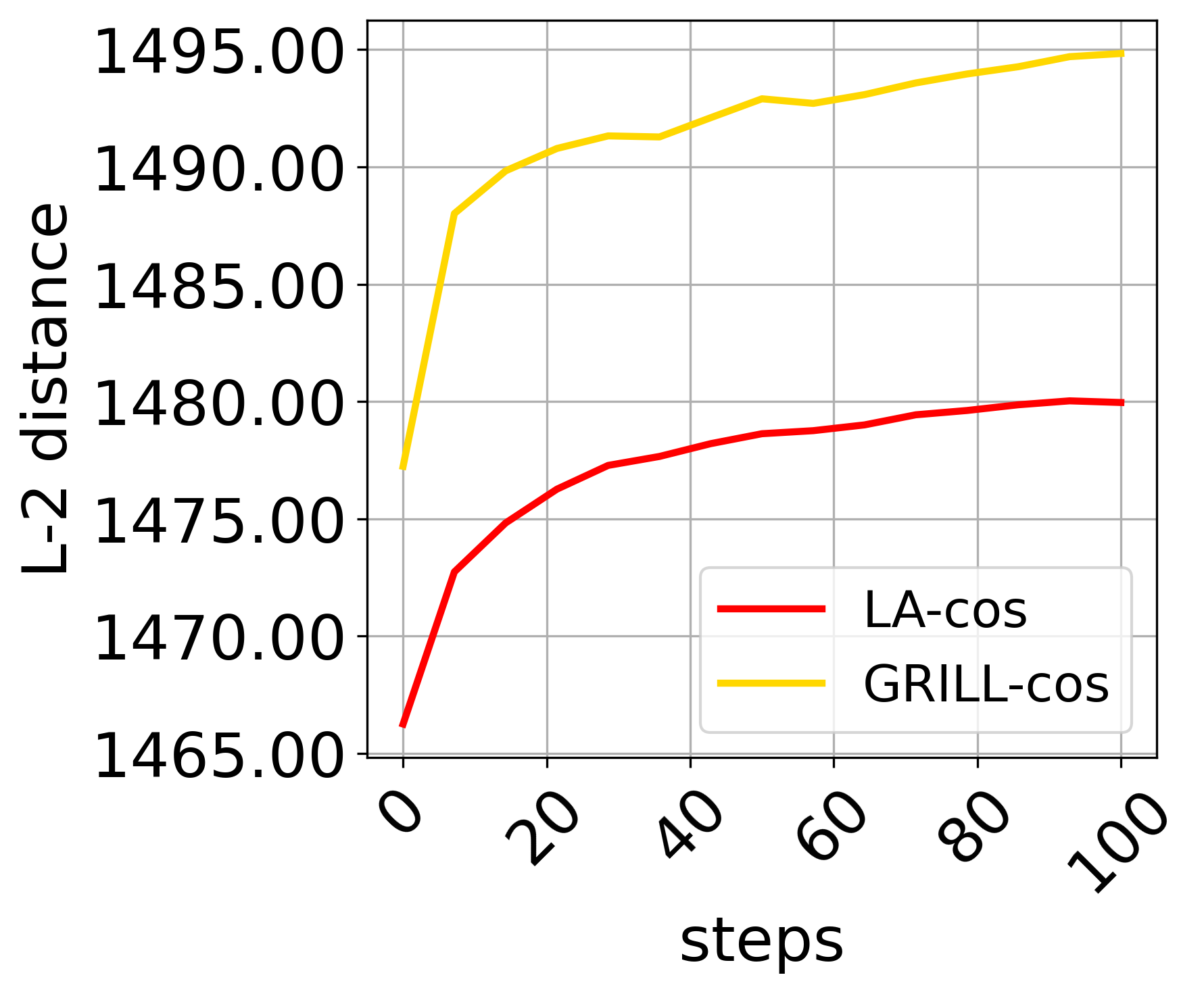}
        \caption*{(b) DiffAE}
    \end{minipage}
    \begin{minipage}[b]{0.32\linewidth}
        \includegraphics[width=\linewidth]{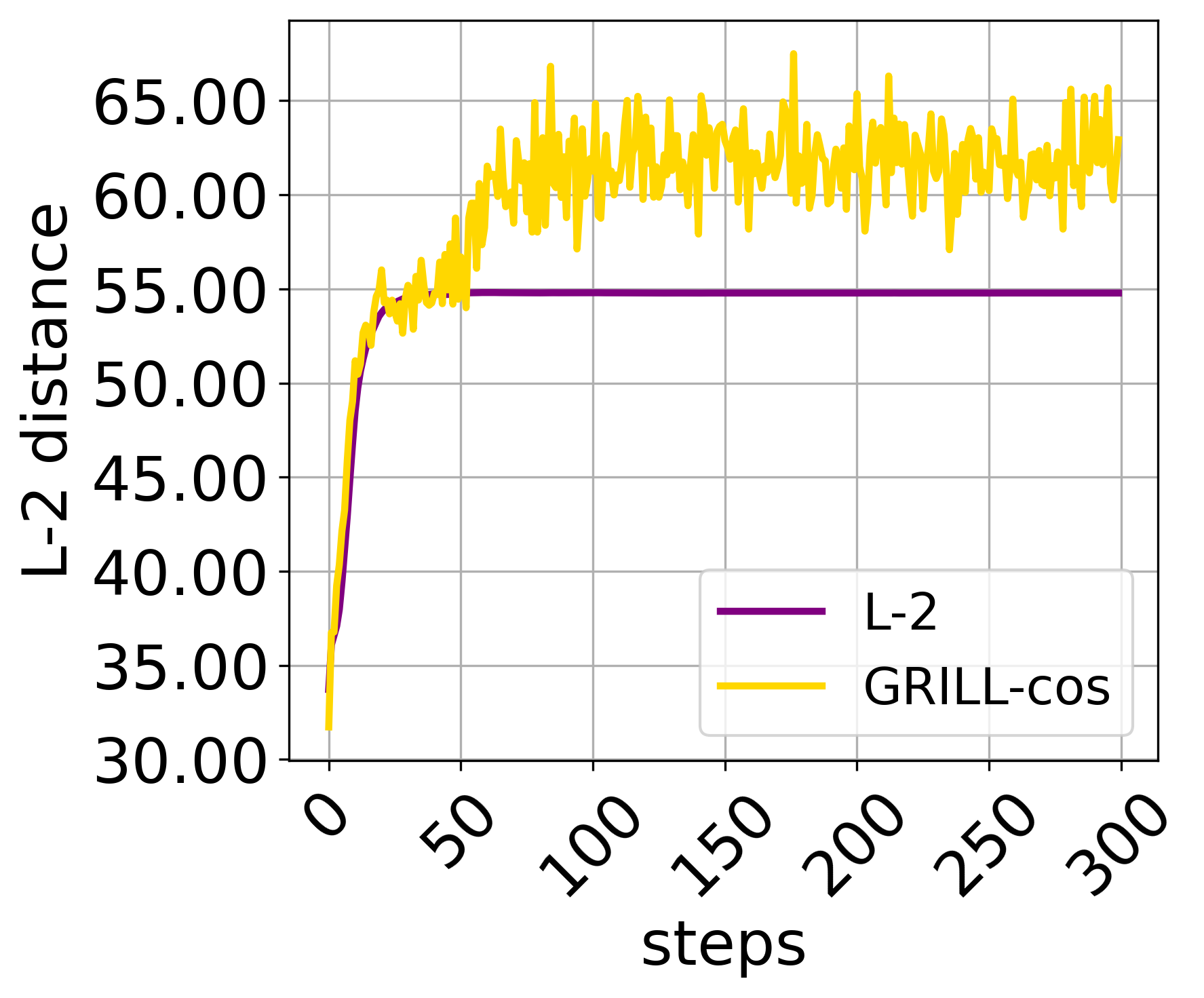}
        \caption*{(c) TC-VAE}
    \end{minipage}
    \begin{minipage}[b]{0.32\linewidth}
        \includegraphics[width=\linewidth]{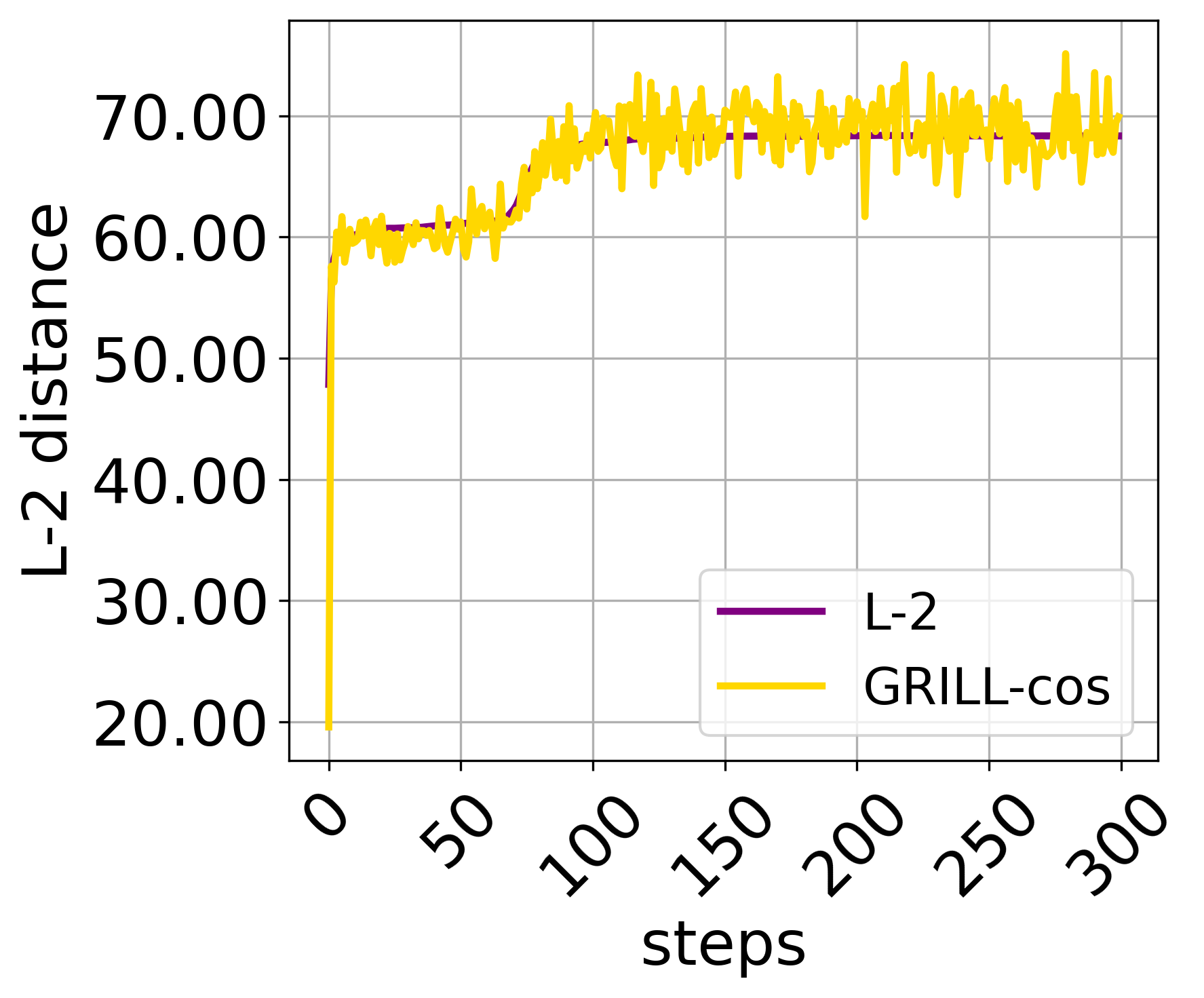}
        \caption*{(d) $\beta$-VAE}
    \end{minipage}
    \begin{minipage}[b]{0.32\linewidth}
        \includegraphics[width=\linewidth]{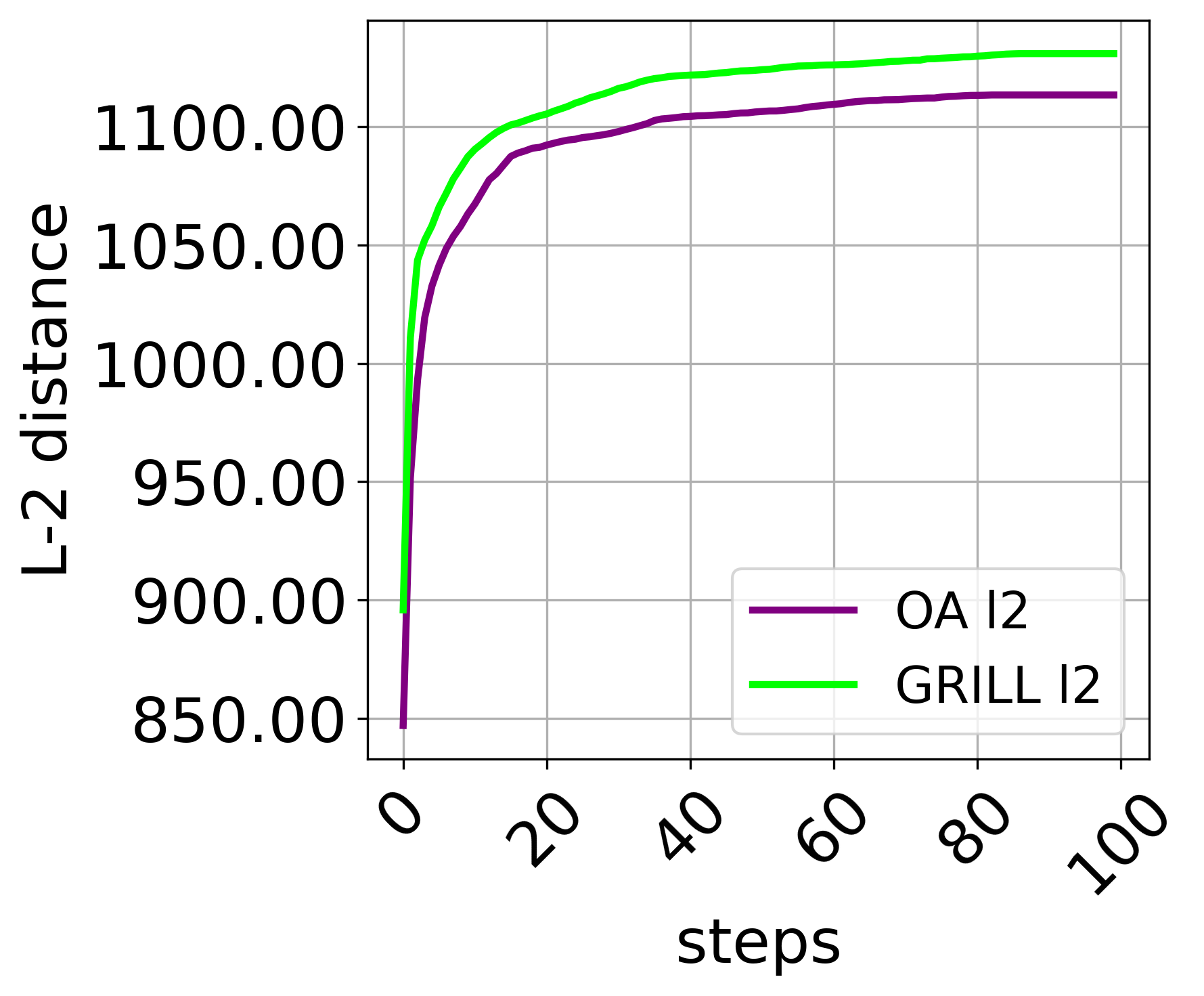}
        \caption*{(e) MAE}
    \end{minipage}
    \caption{Optimization trajectories of \ours{} and the strongest baseline attacks across AEs.}
    \label{fig:appendixConvergence}
\end{figure}

\noindent\textbf{Convergence behavior:} Figure~\ref{fig:appendixConvergence} compares the optimization trajectories of \ours{} and the strongest baseline attacks across all AE architectures. In highly ill-conditioned models such as NVAE and DiffAE, \ours{} converges to substantially higher distortion levels, whereas baseline attacks plateau earlier due to degraded gradients. In comparatively well-conditioned models such as MAE and \(\beta\)-VAE, both methods exhibit similar convergence behavior.

\noindent\textbf{Efficiency:} Table~\ref{tab:combined_flops} reports the floating-point operations (FLOPs) per attack step for AE and VLM attacks. By reusing intermediate representations from the forward/backward pass, \ours{} remains computationally comparable to existing latent-space attacks on AEs and competitive with state-of-the-art VLM attack baselines.

\begin{table}[t]
\centering
\caption{FLOPs per attack step for AEs and VLMs}
\label{tab:combined_flops}
\small
\begin{tabular}{lcc}
\hline
\multicolumn{3}{c}{\textbf{AEs}} \\
\hline
\textbf{Method} & \textbf{NVAE} & \textbf{DiffAE} \\
\hline
LA-(L2, wass, cos)      & $1.73 \times 10^{11}$ & $1.40 \times 10^{11}$ \\
GRILL-(L2, wass, cos)   & $1.73 \times 10^{11}$ & $1.40 \times 10^{11}$ \\
\hline
\\[-0.8em]
\hline
\multicolumn{3}{c}{\textbf{VLMs}} \\
\hline
\textbf{Method} & \textbf{Qwen 2.5} & \textbf{Gemma 3} \\
\hline
BSA         & $2.94 \times 10^{13}$ & $2.59 \times 10^{13}$ \\
GRILL-cos   & $1.70 \times 10^{13}$ & $2.59 \times 10^{13}$ \\
DRA         & $1.42 \times 10^{13}$ & $1.01 \times 10^{13}$ \\
EGA         & $2.68 \times 10^{13}$ & $1.95 \times 10^{13}$ \\
FDA         & $6.14 \times 10^{12}$ & $1.01 \times 10^{13}$ \\
SSPA         & $6.14 \times 10^{12}$ & $1.56 \times 10^{13}$ \\
\hline
\end{tabular}
\end{table}

\section{Conclusions and Outlook}
\label{sec:conclusions}
In this work, we show that ill-conditioned layers with near-zero singular values can suppress adversarial gradient propagation in encoder--decoder architectures, leading to weak norm-bounded attacks and overstated robustness.
We propose \ours{}, a gradient-guided adversarial optimization framework designed to mitigate gradient degradation and consistently improve attack effectiveness across diverse AE architectures. We further present results on vision--language models, 
providing preliminary evidence that related optimization phenomena may also arise in broader encoder–decoder architectures. These findings highlight the importance of conditioning-aware adversarial optimization for reliable robustness evaluation and motivate future research on computationally efficient adversarial evaluation methods for generative and multimodal encoder-decoder systems.

\noindent\textbf{Limitations.} \ours{} is limited to  white-box robustness evaluation in encoder–decoder architectures with continuous latent spaces, and does not directly extend to discrete latent spaces (e.g., VQ-VAE).

\bibliographystyle{ACM-Reference-Format}
\bibliography{main}


\end{document}